\documentclass[10pt]{article} 

\usepackage{listings}
\usepackage{graphicx}

\usepackage{amsmath,amssymb}

\usepackage{changepage}

\usepackage{textcomp,marvosym}

\usepackage{nameref,hyperref}
\hypersetup{
    pdfborder={0 0 0},  
    colorlinks=true,     
    linkcolor=black,      
    citecolor=black,     
    urlcolor=black         
}
\usepackage{lineno}

\usepackage[super]{natbib}

\usepackage{array}
\usepackage{csquotes}

\usepackage{amsmath,amsfonts,bm}

\def\eqref#1{equation~\ref{#1}}

\def\1{\bm{1}}

\def\vp{{\bm{p}}}

\def\evp{{p}}

\DeclareMathAlphabet{\mathsfit}{\encodingdefault}{\sfdefault}{m}{sl}
\SetMathAlphabet{\mathsfit}{bold}{\encodingdefault}{\sfdefault}{bx}{n}

\DeclareMathOperator*{\argmax}{arg\,max}

\title{Foveated Retinotopy Improves Classification and Localization in Convolutional Neural Networks}

\newcommand{\instone}{Institut de Neurosciences de la Timone, Aix-Marseille Université - CNRS UMR 7289, Marseille, France}
\newcommand{\insttwo}{{\'E}cole Centrale Méditerranée, Marseille, France}

\author{
  Jean-Nicolas Jérémie\textsuperscript{1},
  Emmanuel Daucé\textsuperscript{1,2},
  Laurent U Perrinet\textsuperscript{1}\thanks{laurent.perrinet@univ-amu.fr} 
}

\date{} 

\begin{document}

%
\maketitle
\textsuperscript{1}\instone , 
\textsuperscript{2}\insttwo

\section*{Abstract}

From falcons spotting preys to humans recognizing faces, rapid visual abilities depend on a foveated retinal organization which delivers high‑acuity central vision while preserving low‑resolution periphery. This organization is conserved along early visual pathways but remains underexplored in machine learning. Here we examine how embedding a foveated retinotopic transformation as a preprocessing layer impacts convolutional neural networks (CNNs) for image classification. By applying a log‑polar mapping to off‑the‑shelf models and retraining them, we retain comparable accuracy while improving robustness to scale and rotation. We show that this architecture becomes highly sensitive to fixation‑point shifts, and that this sensitivity yields a proxy for defining saliency maps that effectively facilitates object localization. Our results show that foveated retinotopy encodes prior geometric knowledge, offering a solution to visual‑search and enhancing both classification and localization. These findings connect biological vision principles with artificial networks, pointing to new, robust and efficient directions for computer‑vision systems.

\textbf{Keywords:} 
Foveated vision; Convolutional Neural Networks; Transfer learning; Visual categorisation; Neuromorphic transformation; NeuroAI

\section*{Introduction}
\section*{Introduction: Properties of the visuo-motor system endowing visual search}
For predators like birds of prey~\citep{Mitkus2018} or sharks~\citep{Collin2018}, efficiently detecting prey is critical for survival. More broadly, visual search —the process by which organisms scan their environment to locate and identify objects of interest— is essential across species. This includes predators like falcons and sharks, as well as herbivores such as howler monkeys, fruit bats, hummingbirds, and parrots. Many of these species share a visual system characterized by foveated retinotopy~\citep{Polyak1941}, where the visual field is represented more densely around a central region. This organization is believed to underpin the efficiency of visual search.
 

\begin{figure}
   \begin{center}
      \includegraphics[width=.99\textwidth]{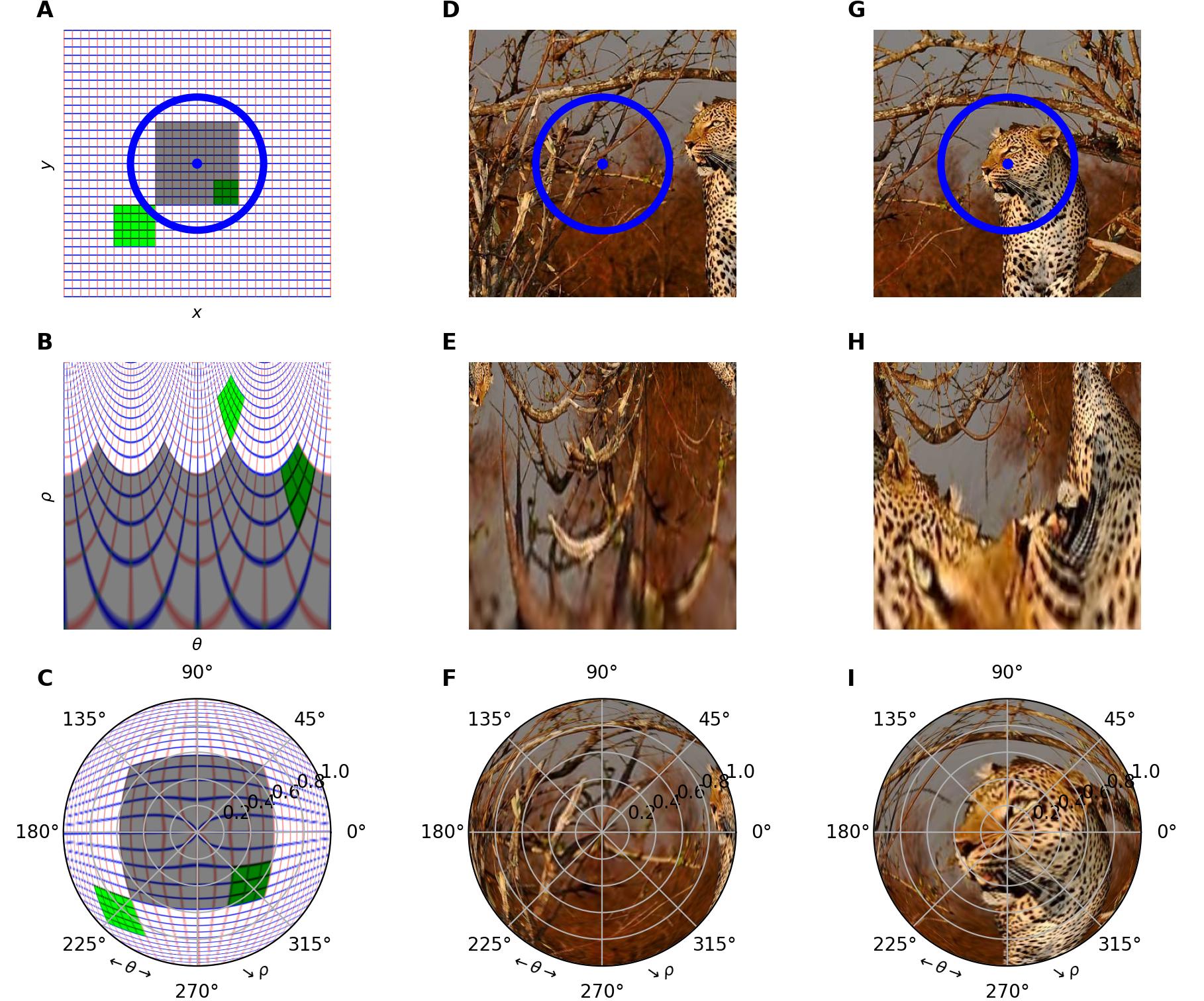} 
   \end{center}
   \caption{
      We illustrate the process of mapping input images defined in Cartesian coordinates to a foveated retinotopic space using a log-polar transformation. The fixation point is marked by a blue disk and the approximate area of the fovea by a blue circle. In \textbf{(A)}, the input image is defined as a regular grid representing the Cartesian coordinates ($x$, $y$) by vertical (red) and horizontal (blue) lines. As shown in \textbf{(B)}, by applying the log-polar transform to this image, the coordinates of each pixel with respect to the fixation point are transformed based on its azimuth angle $\theta$ (abscissa) and the logarithm of its eccentricity $\rho = \log(\sqrt{x^2+y^2})$ (ordinates). This transformation results in a fine-grained representation of the central area and a deformation of the visual space. Note that the green square is translated in retinotopic space when it is scaled and rotated. The third row \textbf{(C, F, I)} illustrates the reconstruction of the image in question, exhibiting a over-representation around the point of fixation in the Cartesian reference frame. When the transformation is applied to a natural image, as shown in \textbf{(D, G)}, there is a noticeable compression of information in the periphery in the log-polar referential (see \textbf{(E, F, H, I)}). Also, this representation is highly dependent on the fixation point, as indicated by the shift shown in \textbf{(G, H, I)} when the fixation point is moved to the right and up.
   } 
\label{logpolar}
\end{figure}

\subsection*{Foveated Retinotopy: A non-linear, radial organization}

While many species, such as prey animals like rabbits, have a uniform retinal topography that assigns roughly equal importance to all parts of the visual field, humans and many mammals possess a foveated retinotopy. This organization features a sharp disparity in resolution, with a high-acuity central area and lower-resolution peripheral regions. Its center defines the point of fixation. Already at the retina, the mapping is mostly radial and the density of photoreceptors decreases exponentially with eccentricity~\citep{Anstis1974}, defining a so called \emph{log-polar} mapping. This radial organization of the retina, with highest acuity in the center and decreasing acuity in the periphery, is largely maintained across the different stages of visual processing in the brain, as shown in the human cortex by functional magnetic resonance imaging (fMRI) to map visual responses~\citep{Tootell1998,Dougherty2003}. A first study by~\citet{Tootell1998} focused on the phenomenon of spatial attention and sought to elucidate the response exhibited when a bar stimulus is presented in a subdivided visual field. As a result, a foveated retinotopic mapping of the visual field is observed in the early visual areas, specifically V1, V2, and V3. \citet{Dougherty2003} then worked out the representation in the same areas when stimuli varying in eccentricity or angular direction were projected in direct association with the log-polar reference. These studies showed how different regions of visual cortex are activated depending on the position of stimuli in the visual field, confirming previously proposed models. 

The precise evolutionary advantage of this foveal organization is still under debate. For example, it is thought to facilitate efficient parallel processing of spatial features in both primates~\citep{nauhaus_efficient_2016} and humans~\citep{golomb_native_2008}. It may also facilitate connections within and between processing areas that respect the geometry of the sensory epithelium and minimize global wiring length~\citep{Lu2023}. However, it has also been suggested that this is merely an artifact of the scaffolding that operates during development~\citep{Weinberg1997}. Another hypothesis is that a fovea implements prior knowledge of visual function with respect to the species' range of behaviors, as it allows efficient use of resources by prioritizing high-resolution vision in areas where it matters most (i.e., the center of gaze) while sacrificing some resolution in peripheral vision. Eagles or bottlenose dolphins are remarkable in this regard because they have a dual fovea, one for fine lateral vision and one for high acuity straight ahead vision. This suggests that each retinal topography provides an efficient solution to the visual search problem for the given species within its respective ecological niche.  More generally, this raises the problem of how the visual system builds an egocentric model of local space from the parcellated retinal input~\cite{Martins2024}.
\subsection*{Eye movements and the sequential analysis of the visual scene}

A natural question is what computational advantages these retinotopic visual inputs confer on information processing. Numerous hypotheses have been proposed regarding the role of this nonuniform visual field mapping. One primary explanation is that a foveal input facilitates visual exploration: a retina with a fovea allows efficient visual processing if the eye can actively move and focus attention on specific points of interest. More generally, the fovea is associated with a set of oculomotor behaviors aimed at positioning objects of interest at the center of the retina (saccades, smooth pursuit, vestibulo-ocular reflex, ...), thus maximizing access to visual information for these objects. In particular, when we visually explore a scene, our eyes perform saccades, rapid eye movements that move the fovea to successive points of interest. In humans, 2 to 3 times per second on average, corresponding to about $150,000$ movements per day and the typical duration of a saccade is very short, about $200$ milliseconds. In between these saccades, fixations allow precise processing of visual information in a short period of time averaging 200 to 300 milliseconds. This alternation between periods of saccades and fixations forms the basis of our scan path and influences how we perceive and interact with our environment. An interesting hypothesis is that the distribution of photoreceptors in foveated retinotopy reflects the probability that an object of interest is located at the center of the retina, given the dynamic parameters of the oculomotor system~\citep{lewis2003distribution,lewis2004understanding}, thus allowing efficient visual exploration.

To better understand how visual search works in humans, it is necessary to study the underlying mechanisms that control our eye movements and visual attention. It is influenced by several factors, including the saliency of objects~\citep{itti_koch2005}. The systematic study of visual search began with the pioneering work of~\citet{Yarbus1961} and~\citet{noton_scanpaths_1971}, who introduced the concept of a ``scan path'' as the trajectory of eye movements produced during visual exploration. In the original experiment conducted by Yarbus, participants were presented with a series of different exploration targets applied to realistic paintings or photographs. The results showed that the scan paths that participants followed were not random, but had a degree of structure that was consistent with their individual goals. \citet{noton_scanpaths_1971} further improved the methods for recording and analyzing these movements by developing an approach that involves non-invasive monocular measurement of eye movements using the diffuse scleral reflection technique, allowing ecological observation of eye movements during recognition tasks~\citep{noton_scanpaths_1971}. Since then, eye-tracking studies have shown that our eyes follow predictable patterns that maximize the efficiency of visual information acquisition~\citep{Rothkopf2016}. Recent studies have shown that this combination of saccades and foveal input, coupled with an effective point-of-interest detection mechanism, significantly enhances visual acuity~\citep{Dauce2020,Dauce2020a,dabane2022you} and supports functions such as the integration of local feature analysis into global perceptual representations. This method provides a promising basis for further investigation of eye movements and the function of foveated retinotopy.
 
\subsection*{The log-polar model in computer vision}

These observations have led to the development of novel models of artificial vision using foveated retinotopy. \citet{Sandini1980} were among the first to develop such models, drawing inspiration from the structures and functions observed in the human eye. In their study, they propose a model that samples the visual scene as a function of eccentricity. It already demonstrates the contribution of this type of transformation to the compression of visual information, emphasizing the visual information in the center and shrinking visual information in the periphery. The most widely used model to represent this topography is the log-polar model, described in detail by~\citet{Araujo1997}. It is organized around two polar coordinates: azimuth (angle) and eccentricity (distance from center) on a logarithmic scale. A log-polar sensor would have a resolution (pixel density) that decays exponentially with distance from the center of the image. Recent developments in neuromorphic computing have also shown that a foveated transformation can be implemented in hardware, allowing real-time processing of visual information~\citep{Faramarzi2024}.

The log-polar transformation has important consequences for the representation of the geometry of objects in the image. It produces important distortions of the original image, with a strong enlargement of the objects present in the center and a shrinking of the peripheral ones (see figure~\ref{logpolar}). In addition, the changes produced by movements of the visual sensor, such as translation, rotation, or zoom, have different consequences in a log-polar reference frame. Any movement of the visual sensor that results in a shift of the center of fixation profoundly alters the distribution of pixels, enlarging the central region and blurring the visual information at the periphery (see Figure~\ref{logpolar}-E\&F). Other camera movements have a more limited effect. In particular, any rotation of the sensor around the central fixation point results in a translation in polar space. Similarly, zooming in or out has the effect of a simple translation in the polar dimension~\citep{Araujo1997}. The log-polar mapping transforms the rotations and zooms into two independent (orthogonal) displacements, since a rotation only changes the azimuth angle $\theta$, while a zoom affects the logarithmic eccentricity $\rho$ on the radial axis. This corresponds to a unique and important property - zooms and rotations in Cartesian space become vertical and horizontal shifts, respectively, in log-polar space (see the green square in Figure~\ref{logpolar}-A,D\&G). 

The log-polar retinotopic mapping was successfully applied in computer vision, especially for template matching~\citep{Araujo1997,Sarvaiya2009,Maiello2020,Palander2008} or robotics~\citep{JavierTraver2010,Antonelli2015}. However, implementing visual processing with retinotopic sensors poses significant challenges in terms of information retrieval: Due to the low peripheral resolution, it is difficult to estimate in advance (before an eye movement is made) which part of the image contains the most relevant information. When analyzing a scene, the eye must infer the regions of interest even before they are positioned at the center of the retina. Two broad categories of approaches can be defined to characterize this anticipatory property of the visual system. The first family of methods (called bottom-up) considers only low-level features to characterize the regions that attract the eye (regions that deviate significantly from the average statistics of the scene). This is the approach proposed, for example, by~\citet{itti_koch2005}. Conversely, top-down approaches use prior knowledge about visual content to direct attention to potentially interesting regions that have not yet been explored~\citep{wolfe1994guided}. Only a few works exploit this principle of ``predictive coding'' in the case of retinotopic sensors; see for example~\citep{najemnik2005optimal, dauce2018active}.

Recent studies have also explored the potential of foveal retinotopy; \citet{Cao2021} designed a new LPNet architecture that includes an internal layer using the log-polar reference. They showed that rotation invariance during categorization can be improved by adding a layer that encodes the input image in the log-polar reference. \citet{da_costa_convolutional_2024} used a CornetZ to investigate the reorganization of the intermediate layer when training the network with retinal ganglion cell (RGC) sampling, a method that produces Cartesian images that mimic foveal retinotopy. It is shown that the use of such inputs has an impact on CNN receptive field mapping. Finally, also using a sampling method, \citet{Lukanov2021} shows that this non-uniform distribution may be beneficial when considering the attentional mechanism for implementing saccades, as it could improve the categorization accuracy of the networks. It seems that the community is increasingly interested in foveal retinotopy and its benefits for the accuracy and robustness of CNNs. These promising results pave the way to a better understanding of this organization, but they overlook an important point: they all rely on a Cartesian reconstruction of the foveal retinal distribution instead of focusing on the log-polar reference frame.  
%
\subsection*{Convolutional neural networks and translational invariance}
%
Deep learning algorithms have made tremendous progress in recent years. For some visual recognition tasks, such as the {\sc ImageNet} challenge~\citep{Russakovsky2015}, convolutional neural networks (CNNs) have made huge strides in computer vision, so much so that they have now surpassed human accuracy in this task~\citep{He2015wrn}. CNNs are particularly well suited to image processing tasks because they allow the operations performed by a neural network to be factorized, while limiting the number of parameters dedicated to each processing layer. Convolutional kernels are adaptive and are trained on many examples for image classification. Although CNNs are trained according to a plasticity rule that is not biologically plausible~\citep{Bengio2016}, they nevertheless have properties that are close to some exhibited by biological visual systems~\citep{Kubilius2018}, in particular (i) massively parallel processing and (ii) the ability to increase the size of the receptive field (the size of a region of the original image) as a function of layer depth. In addition, the ``pooling'' operator also allows a CNN to be less sensitive to the position (or displacement) of objects in the image: we speak of translation invariance, which can be achieved by alternating convolutional layers and pooling layers with increasingly larger receptive fields. CNNs also show striking similarities to the ability of humans to quickly classify images, such as whether an image contains an animal or not~\citep{Jeremie2023}. %

However, one limitation that remains is their vulnerability to adversarial attacks. Studies have shown that these learned models can be fooled by modifications to images that are imperceptible to humans~\citep{Huang2017}. These small distortions cause the algorithms to misclassify examples with high confidence~\citep{szegedy2013intriguing}. This vulnerability makes deep networks unstable and unsuitable for use in safety-critical domains such as medicine, autonomous vehicles, or other life-or-death situations, and requires proper care to generate more robust responses~\citep{Bashivan2023}. Before deep learning can be relied upon for such applications, researchers must find ways to make these models more resilient to adversarial examples and introduce human-level robustness to ensure that mistakes do not have dangerous consequences in the real world. One emerging hypothesis is that insights from biological neuroscience would be critical to achieving this goal~\citep{Mineault2024}.

Despite the growing influence of biological and computational neuroscience on computer vision and machine learning, the contribution of foveated retinotopic mapping to visual processing remains scarce. In particular, it should be highly advantageous to exploit the translational invariance within the feature maps of CNNs as an advantage over the geometry of retinotopic mapping. Indeed, it has been shown that the architecture of any deep learning architecture acts as a ``deep prior'' to constrain the range of achievable representations, so that at one extreme one could use CNNs without learning~\citep{Lempitsky2018}. In this context, the log-polar transformation could be seen as a way to constrain the range of transformations that are possible in the input space, thus providing a more efficient way to learn the underlying structure of the visual world.

Several studies have explored transformations to enhance the robustness of CNNs. For example, \citet{Remmelzwaal2019} demonstrated that log-polar transformations positively impact CNN performance and accuracy in digit discrimination tasks. More recently, \citet{Kim2020} proposed an architecture (CyCNN) that leverages polar coordinate mapping and cylindrical convolutional layers to achieve rotational invariance in image classification tasks, particularly on small datasets. Analogous to these approaches, \citet{esteves_learning_2018} showed that spherical transformations can confer rotational invariance in 3D object categorization. 

Building on these advances, our work extends such frameworks to more complex datasets while requiring only minimal architectural changes, thereby ensuring broad adaptability. In addition, we examine multiple transformations, including rotation, scaling, and translation. Distinct from prior studies, our approach is informed by behavioral neuroscience: we investigate how different reference coordinates shape visual search performance, offering insights at the intersection of biological and computational vision.
\subsection*{Paper contributions}
%
Here, we bridge biological and artificial vision by implementing a log-polar transform which we feed to the input of off-the-shelf deep convolutional neural networks (CNNs), creating a foveated mapping that mimics the spatial organization of biological retinas. Our key contributions are the following.
\begin{itemize}
 \item 
First, despite their effectiveness, we demonstrate that conventional CNNs can exhibit significant vulnerability to basic geometric transformations such as rotations and zooms if not considered correctly, highlighting a fundamental limitation in current architectures.
 \item 
Then, we introduce a biologically-inspired foveated architecture by incorporating a log-polar transformation at the network input, embedding rotation and scale invariance into the model's structure.
 \item 
Next, we validate our approach using transfer learning on standard off-the-shelf architectures and benchmarks, demonstrating that the benefits of foveated vision can be achieved without compromising classification performance when the object is centered.
 \item 
Following that point, we show how our foveated architecture leverages existing convolutional layers in novel ways to enhance robustness against geometric transformations, providing insights into the relationship between network architecture and invariance properties.
 \item 
Moreover, we reveal how the log-polar transformation's sensitivity to fixation point creates an unexpected advantage: by analyzing the probability at the output of the network, we observe variations across multiple viewpoints, such that our system achieves effective object localization without additional training.
 \item 
Finally, we establish connections between computational findings and biological vision, offering new directions for both machine learning architectures and our understanding of natural visual processing. By aligning computational models with neurobiological principles, we aim to improve the performance of artificial systems while also contributing more biologically plausible models to the NeuroAI community.

 \end{itemize}
\section*{Results}
\subsection*{Training on transformed images}
We retrain pre-trained {\sc ResNet}~\citep{He2015wrn} networks on different variants of the {\sc ImageNet}-1K dataset~\citep{Russakovsky2015} (see Methods ``Datasets and Dataset Transformations''), either using a simple circular mask applied on the raw images (hereafter called the Cartesian dataset) or using log-polar transformed images (hereafter called the ``retinotopic'' dataset). Two variants of the training sets are also considered. In a first case (called the ``regular'' case), the mask or the log-polar transformed images are applied to the ``regular'' images. In a second case (called the ``focus'' case), the mask or the log-polar transformation is focused on the center of the \emph{bounding boxes} surrounding the objects of interest and which are provided with the data set. The pre-trained networks are first re-trained on the ``regular'' version of the dataset, generating a first generation of networks, and then a second generation is trained by fine-tuning these networks on the ``focus'' dataset (see Figure~\ref{imagenet_ground_truth}). We then compare the classification accuracy of the original and the different re-trained networks. In this section we choose to express the accuracy as the percentage of correct categorization on the corresponding validation dataset. 

The baseline classification accuracy of the standard pre-trained {\sc ResNet} network, which we will call the ``raw'' network (as it has no mask, retinotopic transformation nor training), averaged on the Cartesian test dataset is $81.7\%$. In comparison, our re-trained networks show respectively accuracies of $78.5\%$ and $74.3\%$ on the ``regular'' Cartesian and retinotopic datasets, and $82.5\%$ and $77.4\%$ on the ``focus'' Cartesian and retinotopic datasets. As we can observe, when fine-tuning the networks on the ``focus'' dataset, despite reducing the image resolution by cropping the image, the accuracy is improved by $4.0\%$ and $3.1\%$ respectively. This demonstrates that this cropping tends to suppress distractors in the periphery and enhances average accuracy. 

Furthermore, our networks re-trained on the log-polar transformed images have a slightly lower categorization accuracy than those re-trained on Cartesian images (see Figure~\ref{rot_accuracy}-A \& Figure~\ref{zoom_accuracy}-A). This result was expected because the log-polar transformation discards fine-grained information in the periphery without increasing central resolution. In fact, this rather limited loss of accuracy is quite remarkable for such a massive loss of information. One reason could be a general ``photographer's bias'' for the ``regular'' images that tends to place the main object in the central region of the image. It is confirmed by the increase of accuracy observed when centering the point of fixation in the ``focus'' dataset. 

\subsection*{Robustness of CNNs}

To what extent can these re-trained networks be relied upon in practical applications? Indeed, a persistent challenge for deep neural networks is their lack of reliability in adversarial situations. For example, it is well documented that a minor alteration to the inputs of these networks, if well-designed, may result in a significant decline in classification accuracy. In particular, a classical robustness test applied to deep networks, called an ``adversarial attack,'' consists of perturbing independently the pixels in each image in order to maximize the error rate in the test phase (see the Methods ``Attacking classical CNNs with a geometrical rotation''). However, these modifications are typically subtle and perceptually resemble identically distributed independent noise, rendering them unlikely to happen in natural conditions. In contrast, biological agents interacting with their environment undergo significant visual perturbations, including large scene pans and tilts due to head and body movements. 

As real-world objects appear in different orientations, we assess the resilience of our re-trained networks to geometric alterations of their inputs e.g. rotations and zooms. In comparison to modifying individual pixels, a rotation or a zoom represents a coherent, whole-image transformation controlled by a single parameter, namely the rotation angle or scaling factor. We thus choose to investigate for each network a simple attack scenario that maximizes the loss for different rotation angles for each image individually and then evaluates the classification result for this ``worst'' angle. Then, we compute the accuracy averaged over a sample of $50,000$ images from the {\sc ImageNet} validation dataset.

\begin{figure}[h!]
   \begin{center}
   \includegraphics[width=1\textwidth]{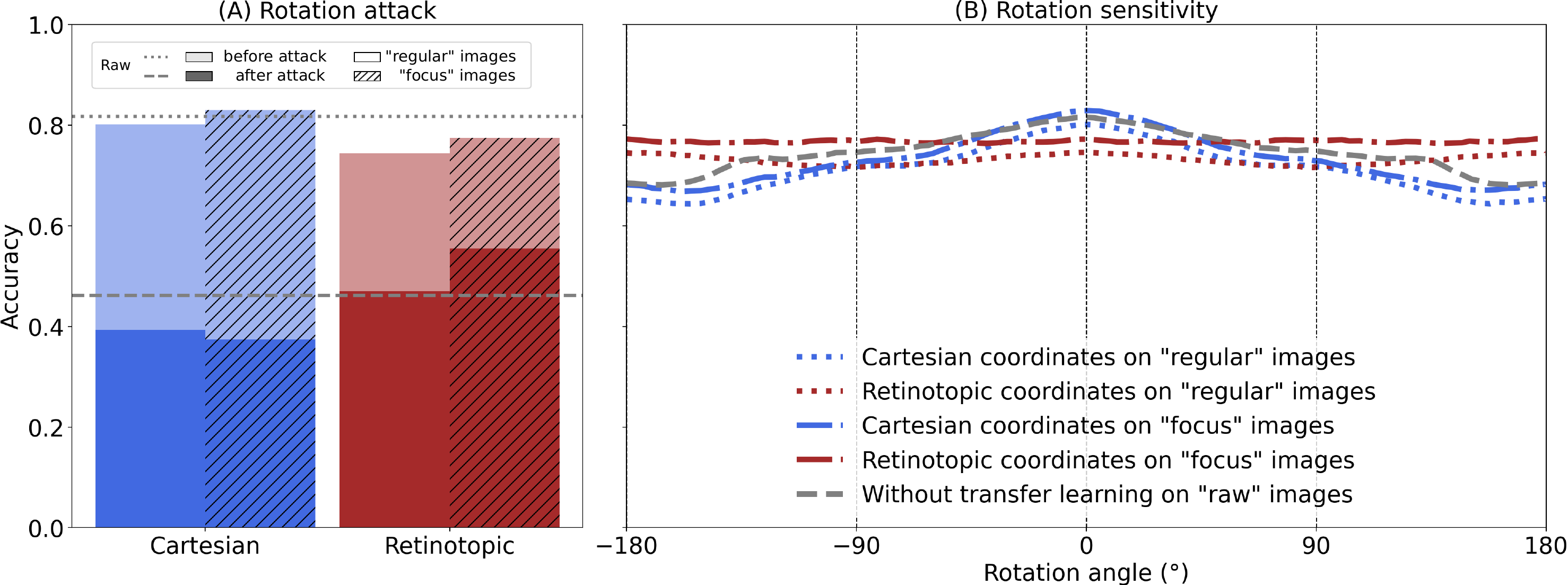}
   \end{center}
   \caption{ \textbf{(A)} For all the networks, we plot the accuracy averaged over the dataset without rotation (in light color), or for each image rotated at the angle $\bar{\theta}$ with the worst loss (rotation-based attack, in full color). No shading : regular dataset. Diagonal shading : focus dataset. Gray dashed lines: accuracy of the ``raw'' network (without any transformation nor training). \textbf{(B)}~The average accuracy is shown for both Cartesian or retinotopic re-trained networks, and the ``raw'' network, with different image rotations. The rotation is applied around the central fixation point with an angle ranging from $-180^\circ$ to $+180^\circ$ (in steps of $15^\circ$).} 
   \label{rot_accuracy}
\end{figure}

Our experiments show that while the original {\sc ResNet} $101$ achieves a nominal baseline accuracy of $81.7\%$ on unperturbed images, a rotation attack significantly reduces the performance of the model. Applying the maximally deceptive rotation to each image reduces the average accuracy to $46.1\%$ (see gray lines in Figure~\ref{rot_accuracy}-A). This is also true for our Cartesian network, with accuracy dropping to $39.3\%$ for the Cartesian network re-trained on ``regular'' images, and even to $37.4\%$ for the network fine-tuned on the ``focus'' dataset. In contrast, the retinotopic networks show a lower sensitivity to rotation attacks, with an accuracy reduced to $47.0\%$ when using the ``regular'' images, and only to $55.4\%$ when using the ``focus'' dataset. 

This difference between the two types of networks is even more manifest in Figure~\ref{rot_accuracy}-B, where we compute the average accuracy over the test dataset for each single rotation angle. We first observe that the Cartesian networks show a decrease in accuracy with respect to the angle of rotation (with a symmetry with respect to horizontal flips), with a significant drop around $160^\circ$, ``raw'' and ``focus'' accuracies were degraded to $66.8\%$ and $68.1\%$, i.e. a drop of $16.1\%$ and $13.4\%$ ($64.3\%$ in ``regular'', i.e. a drop of $15.8\%$). Strikingly, this effect is nearly absent in our retinotopic networks (see Figure~\ref{rot_accuracy}-B), which show a flat (invariant) accuracy over the whole range of rotation angles, with a minimal degraded accuracy at $76.4\%$, i.e. a drop of $0.8\%$ ($71.6\%$ in ``regular'', i.e. a drop of $2.9\%$). This marked difference can be interpreted as a consequence of the horizontal translation invariance found in classical CNNs. When applied to the retinotopic input space, this invariance transforms seamlessly into rotation invariance in visual space~\citep{Araujo1997} (see~\citep{Hao2021} for a proof). Note that for all networks, no maximum accuracy exceeds the value obtained without rotation (i.e. $0^\circ$ rotation angle) by more than $0.01\%$.

\begin{figure}
   \begin{center}
   \includegraphics[width=1\textwidth]{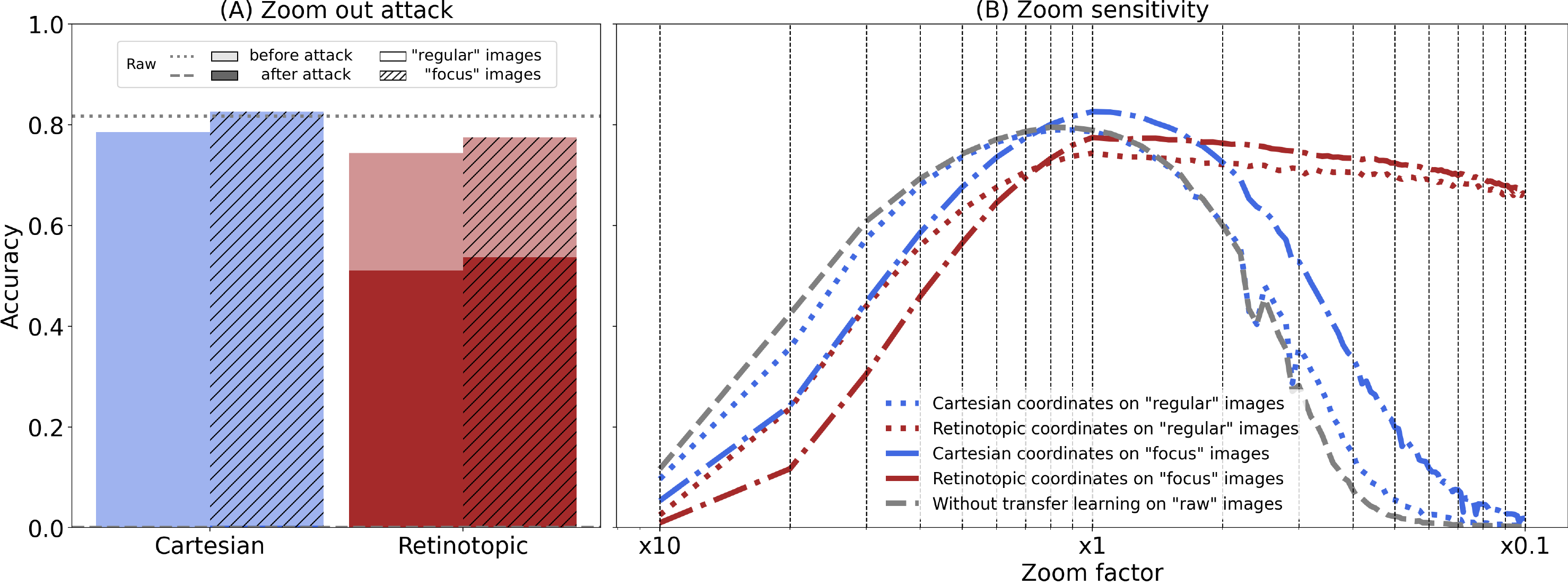}
   \end{center}
   \caption{\textbf{(A)} For all networks, we plot the accuracy averaged over the unzoomed dataset (in light color), or for each image zoomed out to the scale with the worst loss (zoom out attack, in full color). The gray dashed lines represent the accuracy of the ``raw'' network (without re-training). \textbf{(B)}~Average accuracy over a sample of $50,000$ images from the {\sc ImageNet}
   validation dataset, shown for both re-trained and pre-trained networks with different input image zooms. The zoom is applied at the central fixation point with a zoom factor geometrically-spaced from $\times 10$ to $\times 0.1$.}   
   \label{zoom_accuracy}
\end{figure}

Analogous to rotation, zooming in and out is equivalent to a translation in log-polar space, and this property is expected to induce a similar invariance in the retinotopic networks. Similar experiments were therefore performed to test the effect of a zoom (see Figure~\ref{zoom_accuracy}), with a zoom ranging from $\times 10$ to $\times 0.1$, divided into ``zoom-in'' ($\times 10$ to $\times 1$ range) and ``zoom-out'' ($\times 1$ to $\times 0.1$ range). In Figure~\ref{zoom_accuracy}-A, we applied the zoom attack only for the zoom-out case, i.e. ratios between $\times 1$ and $\times 0.1$, because zoom-in attack scenarios always lead to the maximum zoom-in, where the accuracy approaches zero. The  response of all Cartesian networks (i.e. ``raw'', ``regular'' or ``focus'') reach chance-level performance when submitted to a zoom-out attack. 
Only the retinotopic networks keep a discriminative capacity in this case ($51.0\%$ on ``regular'' images and $53.6\%$ on ``focus'' images), illustrating the importance of foveal information in categorization, especially when the peripheral information is scarce or deceptive. 

Figure~\ref{zoom_accuracy}-B illustrates the average accuracy across the entire zoom range, offering a comprehensive perspective on the effects of zooming in and out across our different networks. Unlike the previous case, a pronounced asymmetry emerges in both scenarios. Specifically, the various Cartesian networks tested (both the original and re-trained versions) exhibit an approximate symmetry relative to the logarithmic scale of zoom-in and zoom-out factors. In contrast, the retinotopic network displays an asymmetric response, characterized by a slightly declining plateau in the zoom-out direction and a more pronounced accuracy loss in the zoom-in direction compared to the Cartesian networks. %
When zooming in, both the retinotopic and Cartesian networks exhibit heightened sensitivity, with mean accuracy dropping near chance levels at a $\times 10$ zoom. Notably, the Cartesian networks achieve marginally higher accuracies in this scenario, suggesting a potential architectural advantage under extreme zoom-in conditions. However, this finding should be interpreted cautiously, as it likely stems from the loss of visual detail and the blurring effect inherent to applying extreme zoom levels on low-resolution original images.

In contrast to the case of rotations, our fine-tuning on the ``focus'' dataset has a nuanced effect, as evidenced by its impact on the zoom optimal point. For the Cartesian network, the zoom corresponding to maximum accuracy is around $\times 2$ in both its pre-trained and re-trained states. However, fine-tuning on the ``focus'' dataset shifts this optimal zoom to $\times 1$. In comparison, the optimal zoom for the retinotopic network remains consistently at $\times 1$, regardless of evaluation on ``regular'' or ``focus'' data. From an ecological perspective, zoom-out invariance is likely one of the most advantageous features in natural vision. This capability facilitates the detection of predators or prey at a distance, even in complex and cluttered environments where survival-critical information may appear at varying sizes. While this feature enhances the interpretation of visual scenes, it comes with a trade-off: the necessity to position objects of interest centrally on the retina.

\begin{figure}
   \begin{center}
   \includegraphics[width=1\textwidth]{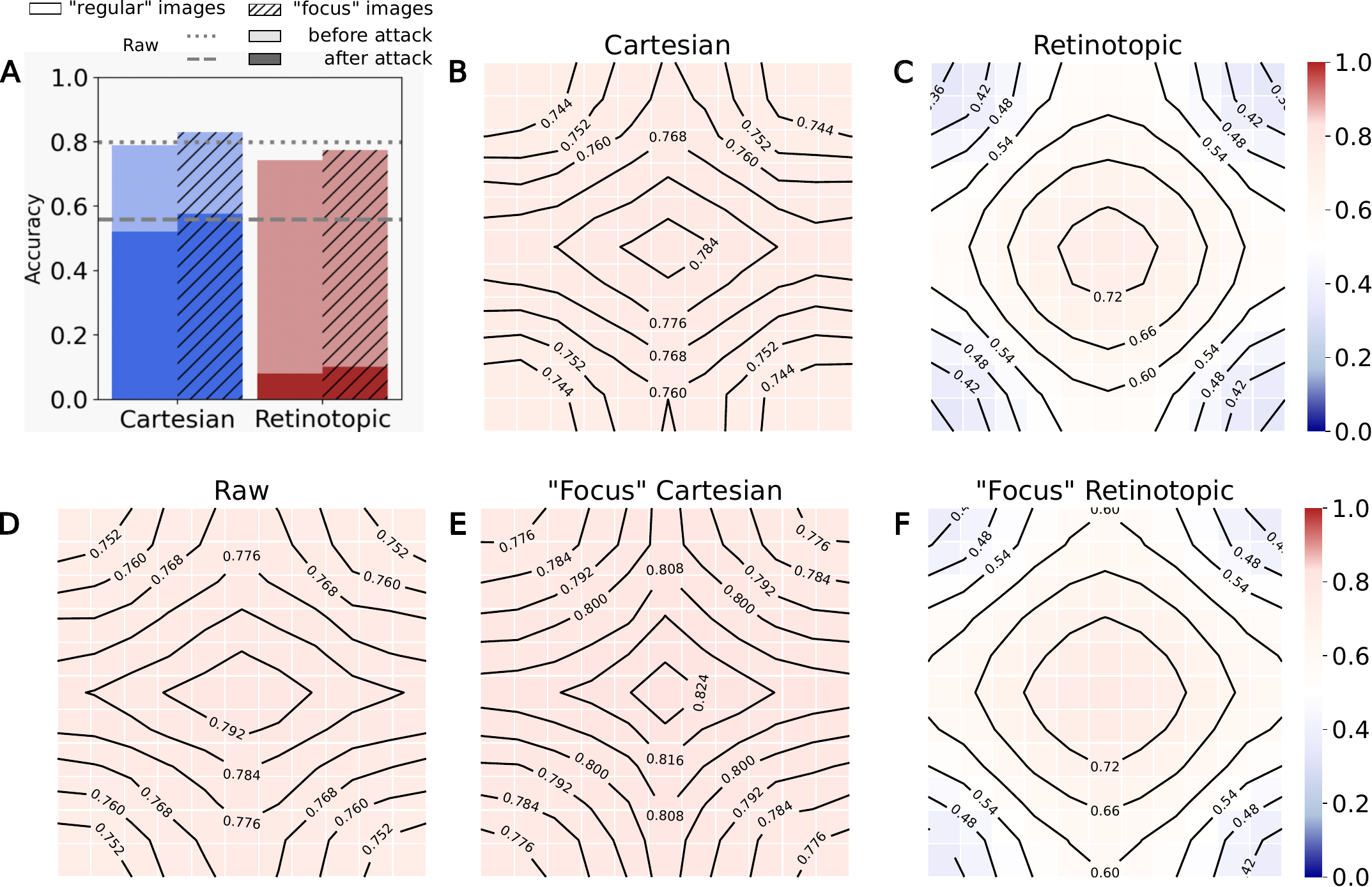}
   \end{center}
   \caption{\textbf{(A)} For all networks, we plot the accuracy averaged over the validation dataset (in light color), or for each image translated (rolled) to the position with the worst loss (translation attack, in full color). The gray dashed lines represent the accuracy of the ``raw'' network (without re-training). \textbf{(B)}~Average accuracy over a sample of $50,000$ images from the {\sc ImageNet} validation dataset, shown for both re-trained and pre-trained networks with different input image translation rolls. The translation is applied from the central fixation point and defines a linear grid of $11 \times 11$ points of fixation distributed regularly on the image.}    
   \label{translation_accuracy}
\end{figure}

Finally, based on these observations and the fact that translations in Cartesian space induce a significant, nonlinear transformation in retinotopic space (see Figure~\ref{logpolar}-A\&B), we investigate the effect of translations in Cartesian space on retinotopic networks. We thus investigate the effect of a rigid full-field translation by applying a roll function to the input image and place the fixation point at different positions in the image. Specifically, the fixation points are linearly distributed on an $11 \times 11$ grid. We then plot the mean accuracy when systematically selecting the worst position (based on the minimum loss as in the scenarios for a rotation Figure~\ref{rot_accuracy}-A or zoom-out attack Figure~\ref{zoom_accuracy}-A) for the target label (see Figure~\ref{translation_accuracy}-A), or the mean accuracy of the networks as a function of the position repositioned in the center (see Figure~\ref{translation_accuracy}-B).

During this attack (see Figure~\ref{translation_accuracy}-A), the accuracy of the retinotopic networks is degraded to $10.1\%$, i.e. a drop of $67.3\%$ ($8\%$ in ``regular'', i.e. a drop of $66.3\%$), while the Cartesian ``raw'' and ``focus'' accuracies were degraded to $55.9\%$ and $57.5\%$, i.e. a drop of $24\%$ and $25.5\%$ ($52\%$ in ``regular'', i.e. a drop of $26.9\%$). As expected, in contrast to previous attacks, the resilience of retinotopic networks is below that of Cartesian networks. Looking at the average accuracy maps (see Figure~\ref{translation_accuracy}-B), we can see that the fixation points around the centre of the map have higher accuracy values than the fixation points around the periphery. This could be an artefact due to the photographer's bias explained earlier. However, this effect is not observed on the Cartesian maps, which shows similar accuracy for all positions examined. Indeed, the robustness of our retinotopic networks to zoom-out and rotation comes at the cost of a high sensitivity to image translations. This increased sensitivity, although detrimental for classification tasks, is associated, as we will see, with an important capability to localize objects of interest in visual space, providing a basis for spatial processing in the brain.

\subsection*{Visual object localization : Likelihood maps}
 \subsubsection*{Protocol} 

\begin{figure}
   \begin{center}
   \includegraphics[width=1\textwidth]{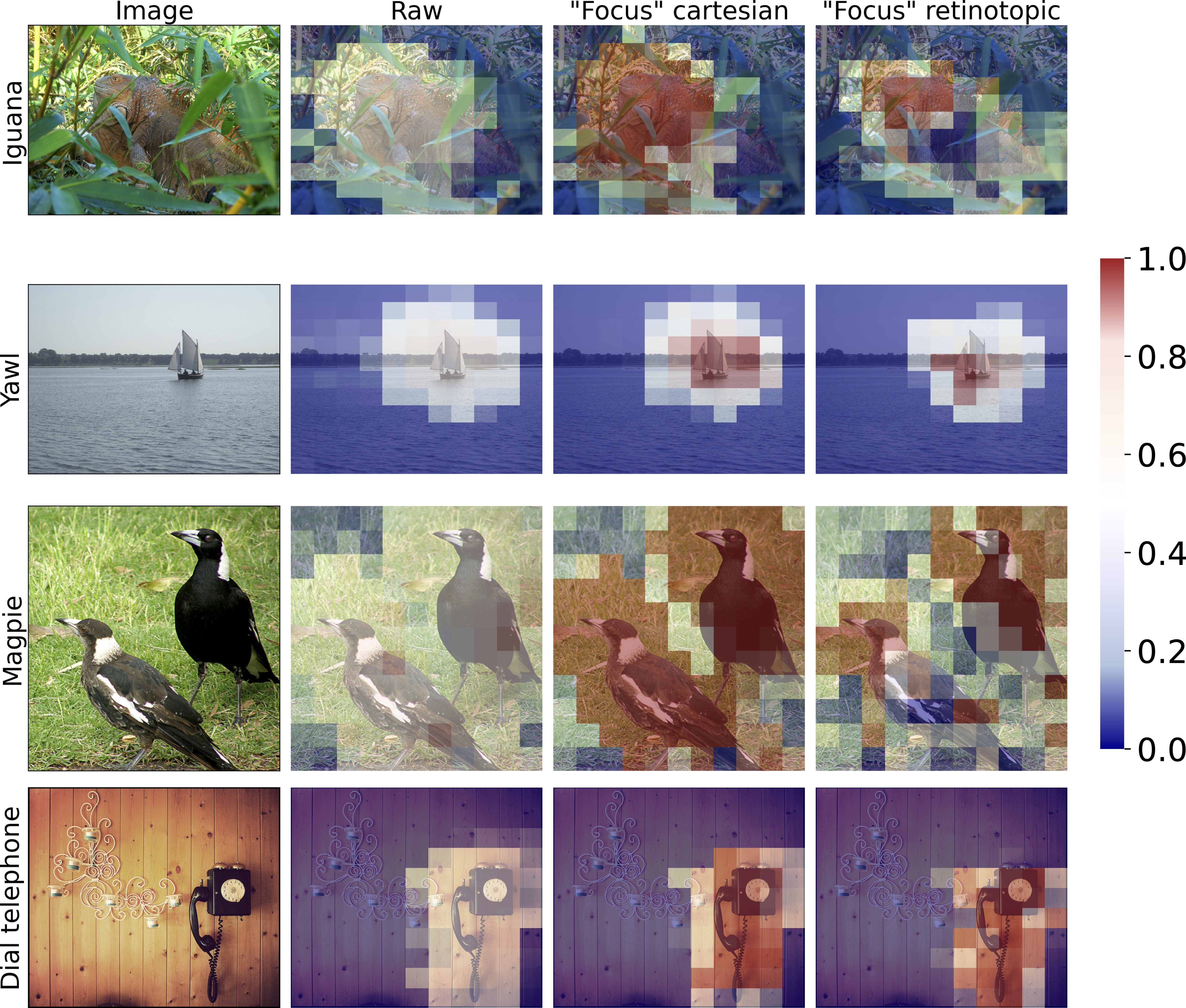} 
   \end{center}
   \caption{Likelihood maps computed on some prototypical images using $11 \times 11$ fixation points with the ``raw'' network, that is the original {\sc ResNet} classifier with no re-training (second column), the network re-trained on images with circular mask on the ``focus'' dataset (third column), and the one re-trained on log-polar tranformed inputs (last column). The map displays the likelihood for the label of interest in the image. 
   } 
   \label{likelihood_maps}
\end{figure}

To quantify the contribution of this sensitivity to translation, we consider here a new task, i.e. the \emph{visual search} task, in which a visual object (of which the label is known in advance) needs to be localized over the entire image. Is is for instance known that such task allows for the quick retrieval of an image label~\citep{Potter1975}. We design a protocol for each network to allow us to compare different visual shifts, each one corresponding to a potential fixation point, and to generate a map of the expected (or actual) accuracy as a function of the fixation point (see Methods ``Visual object localization : Protocol''). In this protocol, a set of $11\times11$ fixation points is defined (with the coordinate $(5,5)$ being the center of the image), and at each coordinate of the grid, a \emph{likelihood} value is computed for the label of interest (see figure \ref{likelihood_map_protocol}). In practice, a likelihood is given at each location from a softmax calculation over the different labels, as in the classical {\sc ResNet} classifier, providing a value between $0$ and $1$ for the label of interest. This projects the network output onto a $11 \times 11$ Bernoulli probability space corresponding to the likelihood of detecting the given label at each position, finally providing a ``heat map'' on our $11\times11$ grid in a way that is compatible with other localization protocols. We tested our likelihood protocol on the ``regular'' validation dataset (see Fig~\ref{likelihood_maps} for some examples).

 \subsubsection*{Mean likelihood maps}

\begin{figure}
   \centering
   \includegraphics[width=1\linewidth]{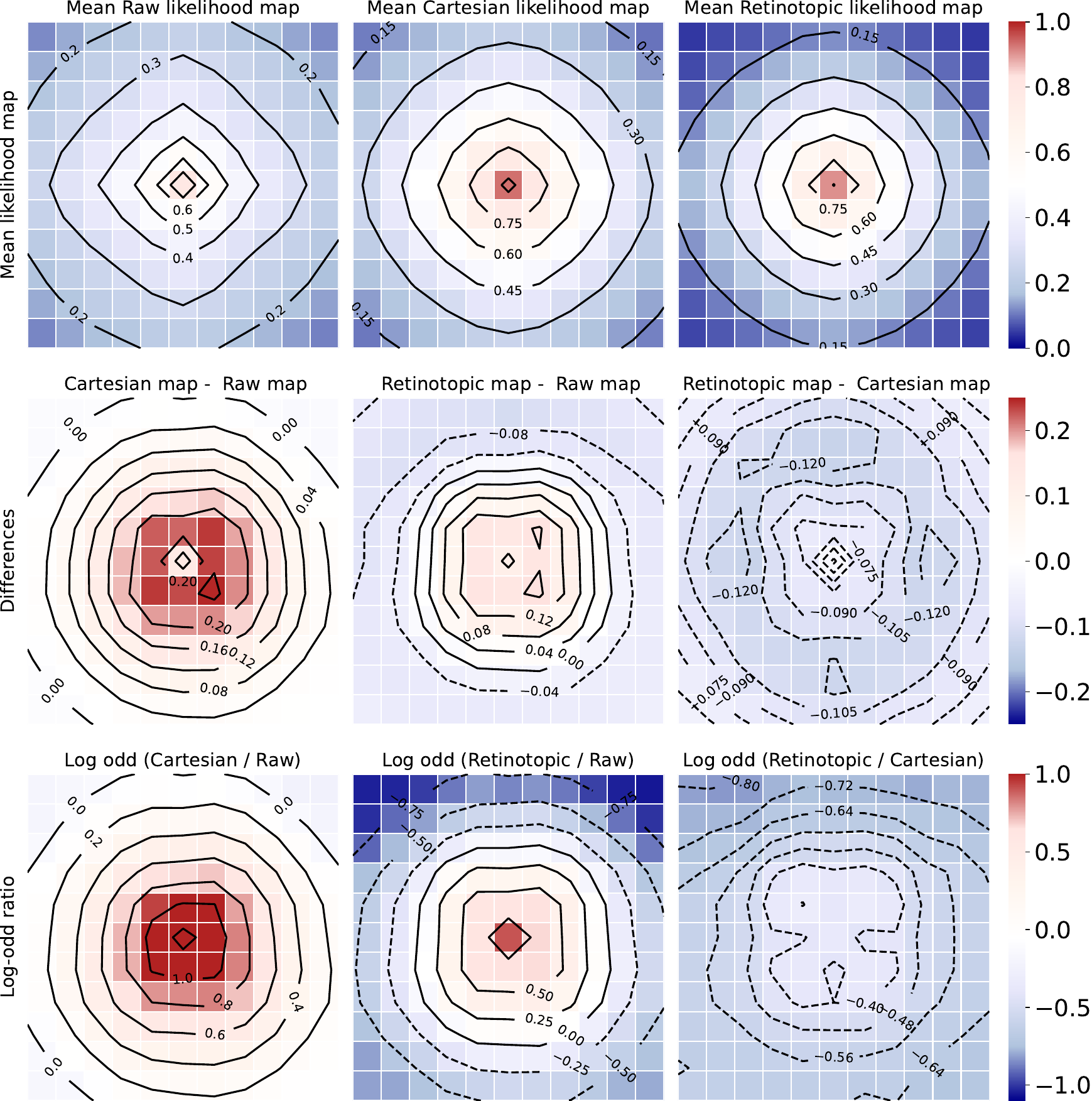}%
   \caption{Mean likelihood on {\sc ImageNet}'s validation dataset ($50 000$ images). From left to right: ``raw'' network (no re-training), Cartesian network retrained on the ``focus'' dataset, and Retinotopic network retrained on the ``focus'' dataset. {\bf First row:} recentered maps, averaged over the validation dataset. {\bf Middle row:} Difference maps. {\bf Bottom row:} Log-odd ratio maps.}
   \label{complete_mean_maps}
\end{figure}

In Figure~\ref{complete_mean_maps}, we calculated the likelihood maps for all the images of the validation set, and re-centered them to place the viewpoint with the highest likelihood at the center of the grid. During the re-centering process, the spots outside the  grid were assigned a \verb|Nan| value to facilitate boundary management, and the average ``center maps'' were generated using the \verb|nanmean()| function,  providing the likelihood profile of the label of interest as a function of the distance from the most salient position. Our recentered likelihood maps are shown on Figure~\ref{complete_mean_maps}. All three maps show a similar 2D bell-shaped activation, indicating a clear object positioning capability for all networks, retinotopic or not, when applying our ``visual search'' protocol (i.e. without rolling the image borders). The activation level is higher at the center (``near'' the object of interest), and lower at the periphery. The likelihood values are different though, being higher on average for the Cartesian network, lower on average for the ``raw'' network, and more contrasted for the retinotopic network. To further quantify this contrast difference, we calculate the map difference on the second row of Figure~\ref{complete_mean_maps}, along with the log-odd ratio in the third row of Figure~\ref{complete_mean_maps}, considering one-to-one comparisons, i.e. Cartesian vs. ``raw'', retinotopic vs. ``raw'', and Cartesian vs. retinotopic.

Let's analyze Figure~\ref{complete_mean_maps}. The second row shows the difference maps. The Cartesian minus ``raw'' map shows only positive values, while the retinotopic minus ``raw'' network has both positive and negative values. The retinotopic minus Cartesian map has only negative values, indicating on average a higher likelihood level in the Cartesian case. The area around the center remains close to zero though, reflecting a sharper slope towards the peripheral region in the retinotopic case. %
The log-odd ratio maps (third row) look quite similar at first sight to the difference maps, except in the center where the differences are more manifest. Of particular interest is the comparison of the first and second column, i.e. Cartesian vs. ``raw'' (first column), and Retinotopic vs. ``raw'' (second column). The radius of the central spot appears clearly different in the two cases, with a large central spot in the Cartesian vs. raw case, reflecting a loosy spatial discrimination improvement, and a more reduced one in the retinotopic vs. raw case, reflecting a more local and object-centric spatial discrimination improvement. Moreover, only the retinotopic network (second column) has significant peripheral depletion in comparison with the ``raw'' network. These observations collectively support the idea that the retinotopic network generates a heat map with enhanced contrast around the area of interest. In addition though, a focal Cartesian network also manifests good object localization capabilities, with high likelihood values around the object of interest, though being less spatially specific.

 \subsubsection*{Comparing likelihood maps and Intersection over Union} 
\begin{figure}
   \centering
   \includegraphics[width=1\linewidth]{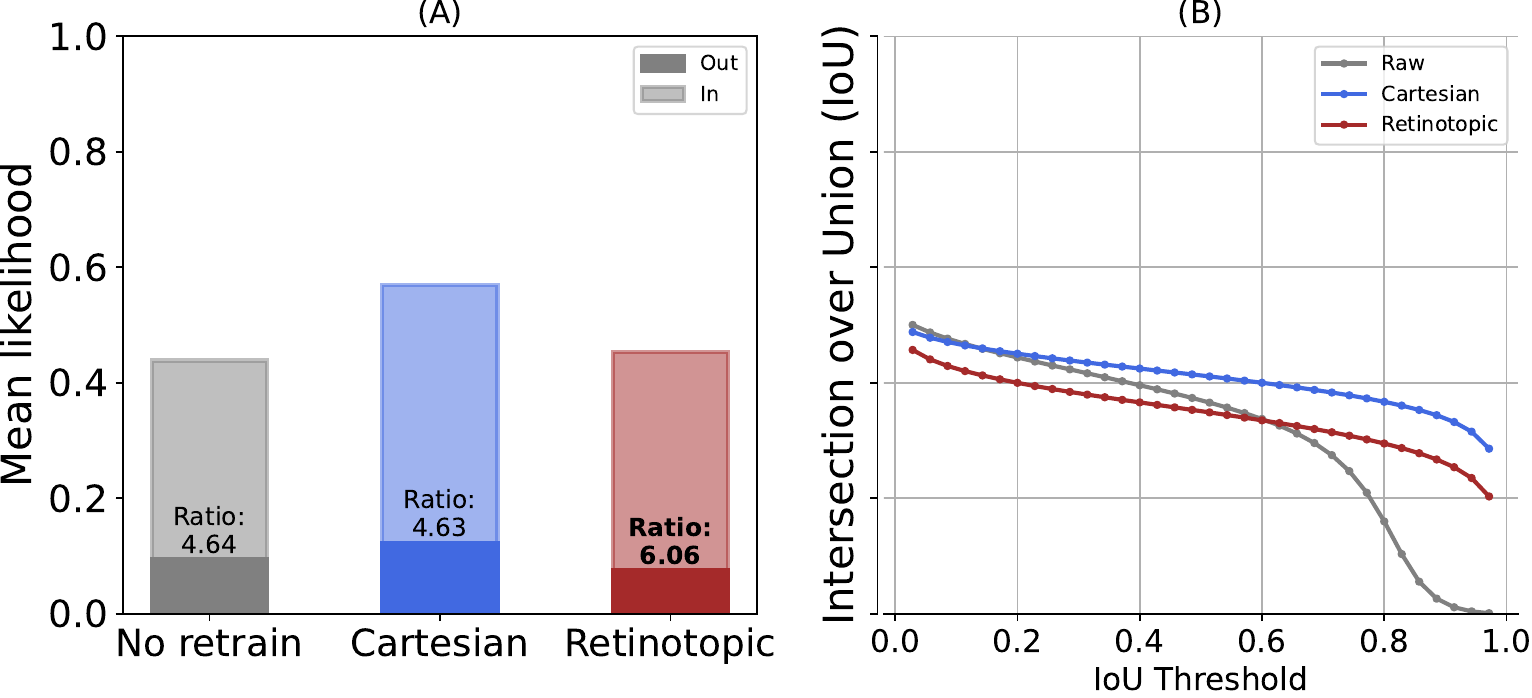}%
   \caption{We tested the visual search protocol on the {\sc ImageNet} validation dataset ($50,000$ images). {\bf (A)}~The mean likelihood across the point of fixation inside the bounding box (``In'') or the point of fixation outside the bounding box (``Out'') and the corresponding ratio of activation. {\bf (B)}~The intersection over union as a function of a threshold applied on the likelihood map.}
   \label{accuracy_activation_Imagenet}
\end{figure}

To further quantify the difference between the Cartesian and retinotopic cases, we considered the localization information provided by the bounding boxes. A bounding box is defined as the rectangle of minimal surface containg one whole visual object. As such, each bounding box partitions the image into two regions, a region where the object of interest is present and a region where it is absent. Thus, given each likelihood map in the validation set, we compute the mean likelihood for the label under study \emph{within} the bounding boxes (ground truth from the dataset) and the mean likelihood \emph{outside} these boxes. Given the higher classification accuracies of the networks when fine-tuned on bounding boxes only the networks retrained on the ``focus'' dataset are used for the remainder of the study. 
The results are shown in Figure~\ref{accuracy_activation_Imagenet}-A. The figure shows a mean likelihood value obtained inside and outside the bounding boxes for the original {\sc ResNet} network, our re-trained Cartesian network, and the re-trained retinotopic network. Both networks show significantly higher likelihoods when the fixation point is inside the bounding box than when it is outside the bounding box. This reflects a higher confidence in the label response. At first glance, the average likelihoods in the three conditions seem quite comparable, although the likelihood values appear slightly higher in the re-trained Cartesian case.

For a quantitative comparison, we then calculate the likelihood ratio between the area inside and outside the bounding box (see Figure~\ref{accuracy_activation_Imagenet}), shown at the top of the bars: the retinotopic network has a higher likelihood ratio than the two Cartesian ones, i.e. $6.1$ versus $4.6$, providing quantitative evidence for a higher contrast of localization in favor of the retinotopic networks. This higher contrast is instrumental in localization tasks, as it allows for better identification of the region of interest. This effect can also be deduced from the examples shown in Figure~\ref{likelihood_maps}: in the retinotopic case, regions of high likelihood are more sparse but still highly contrasted.

In addition, we consider in Figure~\ref{accuracy_activation_Imagenet}-B the Intersection over Union (IoU) metric to evaluate the agreement between the bounding boxes and the activation maps (see Methods ``Localization tools and evaluation''). In contrast with the ``raw'' network, our two networks fine-tuned on the ``focus'' datasets show a slower decay rate, reflecting a better fit with the ground truth. As expected from previous remarks, the IoU stays constistently higher for the cartesian network than for the retinotopic one, illustrating a tendancy for the retinotopic networks to ``concentrate'' likelihoods on smaller portions of the bounding box. The peak IoU is obtained for likelihood thresholds close to $0$ for all networks, this could be an artifact of the measurement, since we are using the bounding box as a comparison, which encourages a larger number of positions with increased likelihood values on the map, rather than a sharp contour around the object of interest. 

\begin{figure}[b]
\begin{center}
  \includegraphics[width=.95\linewidth]{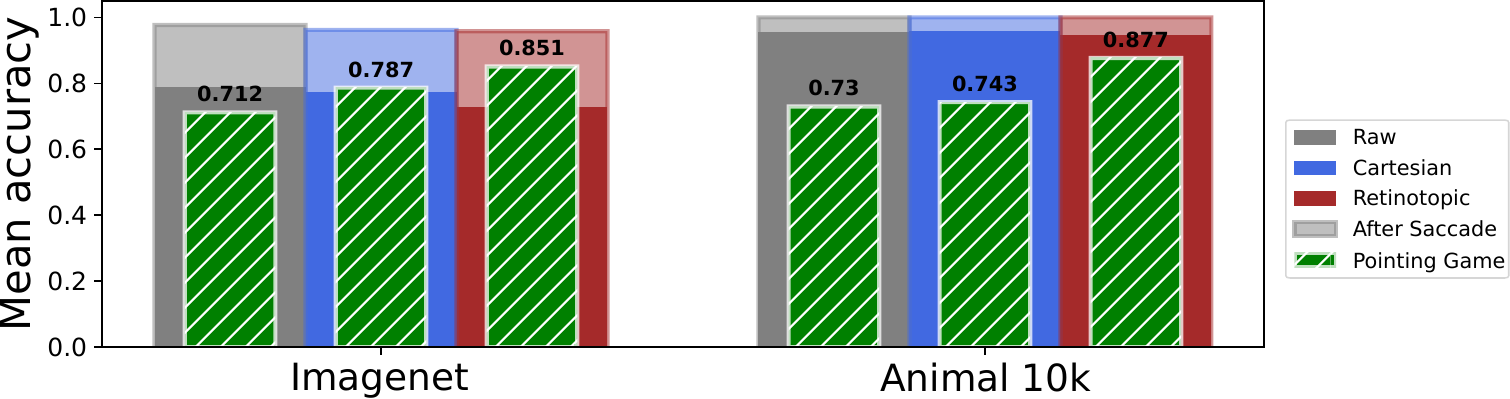}
  \caption{The visual search protocol is tested on the {\sc ImageNet-1K} validation dataset (comprising $50,000$ images) (left) and on the {\sc Animal 10k} dataset (comprising $10,015$ images) (right). The mean accuracy of the networks is displayed, along with the accuracy in the pointing game. }
  \label{pointing_game_accuracy}
\end{center}
\end{figure}

 \subsubsection*{Pointing game}

We have shown that the likelihood maps provide us with an indication, for each image, of the best fixation point to identify the object we are looking for. However, this spatial indication (where to place the eye) does not tell us anything about the visual content of this fixation point. To estimate this visual content more precisely, we now consider the ``pointing game'' metric, that is the rate of successful pointing (i.e. landing \emph{inside} the bounding box) when choosing the highest likelihood position on the likelihood map. The bounding boxes provide us with rather coarse information about the visual content of the image, dividing it into two zones: a box within which the object is present, and a box outline where it is assumed absent. This binary information (``in'' or ``out'') is, however, sufficient to indicate whether the point of highest likelihood is correctly located on or in the immediate vicinity of the object of interest. 

We compute for each network on the validation set the percentage of successful pointings. The results are shown  on Figure~\ref{pointing_game_accuracy} (green bars, left panel), on the raw, Cartesian and Retinotopic networks. The ``raw'' network has $71.2\%$ success, the Cartesian has $78.74\%$ success and the Retinotopic $85.14\%$ success. Those rather high pointing game values reflect the fact that, on average, the bounding boxes cover a large portion of the image in the {\sc Imagenet-1k} dataset. Nevertheless, the retinotopic network shows pointing scores significantly higher than those of the raw network and the Cartesian network. The position of maximal likelihood appears to be strongly correlated with the presence of the object, and this effect is notably sharper and more precise in the case of the retinotopic network.

We extend this visual search strategy by considering the classification accuracy obtained when placing the eye at the position of highest likelihood on the likelihood map, interpreted as a ``saccade'' toward the searched object. By doing so, the accuracy of all networks increases to about $96\%$ (``raw'' $97.7\%$, Cartesian $96.2\%$, retinotopic $95.8\%$, see Figure~\ref{pointing_game_accuracy}), providing an estimation of the \emph{improvement} that can be expected, with respect to the baseline, when placing the eye (or camera) at its highest likelihood position (interpreted as the highest \emph{likely} position). Consistently with our previous results, the accuracy improvement is higher in the retinotopic case, with $22.9\%$ improvement, to be compared with $18.7\%$  and $18.9\%$ improvement in the ``raw'' and Cartesian cases respectively (see Figure~\ref{pointing_game_accuracy}). This strong response improvement reflects the critical role of the ``direction of sight'' in image classification, even in the Cartesian case, an issue little evoked in mainstream computer vision. It also suggest that the retinotopic network could be further improved by introducing such saccades during the retraining phase.

 \subsubsection*{Impact of network depth on localization and classification}

\begin{table} 
   \caption{Analyses of the key metrics for localisation (in/out likelihood ratio and pointing game) and classification (central fixation and max likelihood (``saccade'') fixation) for different network depths.}
   \label{network_depth}
  \begin{center}
       \resizebox{.95\textwidth}{!}{ %
          \begin{tabular}{c||c|c||c|c||c|c||}
          \hline
           & \multicolumn{2}{|c||}{{\sc ResNet} $18$ }& \multicolumn{2}{|c||}{{\sc ResNet} $50$ }& \multicolumn{2}{|c||}{{\sc ResNet} $101$}\\
           \hline
               & \multicolumn{1}{|c|}{Cartesian} & \multicolumn{1}{|c||}{Retinotopic} 
              & \multicolumn{1}{|c|}{Cartesian} & \multicolumn{1}{|c||}{Retinotopic} 
              & \multicolumn{1}{|c|}{Cartesian}  & \multicolumn{1}{|c||}{Retinotopic} \\
          \hline
          \hline
          In/Out Likelihood Ratio &  6.50 &  {\bf 9.02} &  5.02 &  {\bf 6.40} &   4.63 &  {\bf 6.06}\\
          \hline
          Pointing Game  & 83.07\% & {\bf 87.40\%} & 80.81\% & {\bf 85.13\%} &  78.74\% & {\bf 85.14\%}\\
          \hline
          Central fixation & {\bf 0.62 } &  0.52 & {\bf 0.76} & 0.71 &  {\bf 0.77 } &   0.73\\
          \hline
          With saccade & {\bf 0.92 } & 0.86  & {\bf 0.95} &  0.94  &  {\bf 0.96} &  0.95\\
          \hline
          Accuracy increase  & +30\% & {\bf +34\% } & +21\% & {\bf +23\%} &  +19\% &  {\bf +22\%}\\
          \hline
       \end{tabular}
  }
  \end{center}
\end{table}  

We extend the analysis by considering network depths, ranging from {\sc ResNet}~$18$ to {\sc ResNet}~$101$. The results are shown in Table~\ref{network_depth}. We observe the same general trend across different network sizes: classification accuracy improves when the fixation point is placed at the position of highest likelihood, likelihood values are higher inside the bounding box, and the fixation point is predominantly located within the bounding box. These trends are present in both types of networks (Cartesian or Retinotopic), but they are more pronounced in the retinotopic networks. 
Network size primarily affects the classification rate, with lower rates observed for smaller networks. Classification rates remain slightly higher for Cartesian networks in all cases, but the improvement in classification rate based on fixation position is consistently stronger for retinotopic networks. Regarding localization (likelihood ratio and pointing game), network size does not appear to be a determining factor, with higher pointing scores and likelihood ratios observed for smaller networks, and again, a systematic advantage for retinotopic networks.
In summary, Cartesian networks maintain a slight advantage in categorization, while retinotopic networks show improved performance in localization. It is noteworthy that smaller networks demonstrate better localization capabilities, challenging the conventional wisdom that performance necessarily increases with the number of layers.

\begin{figure}[!ht]

   \begin{center}
   
         \includegraphics[width=1\linewidth]{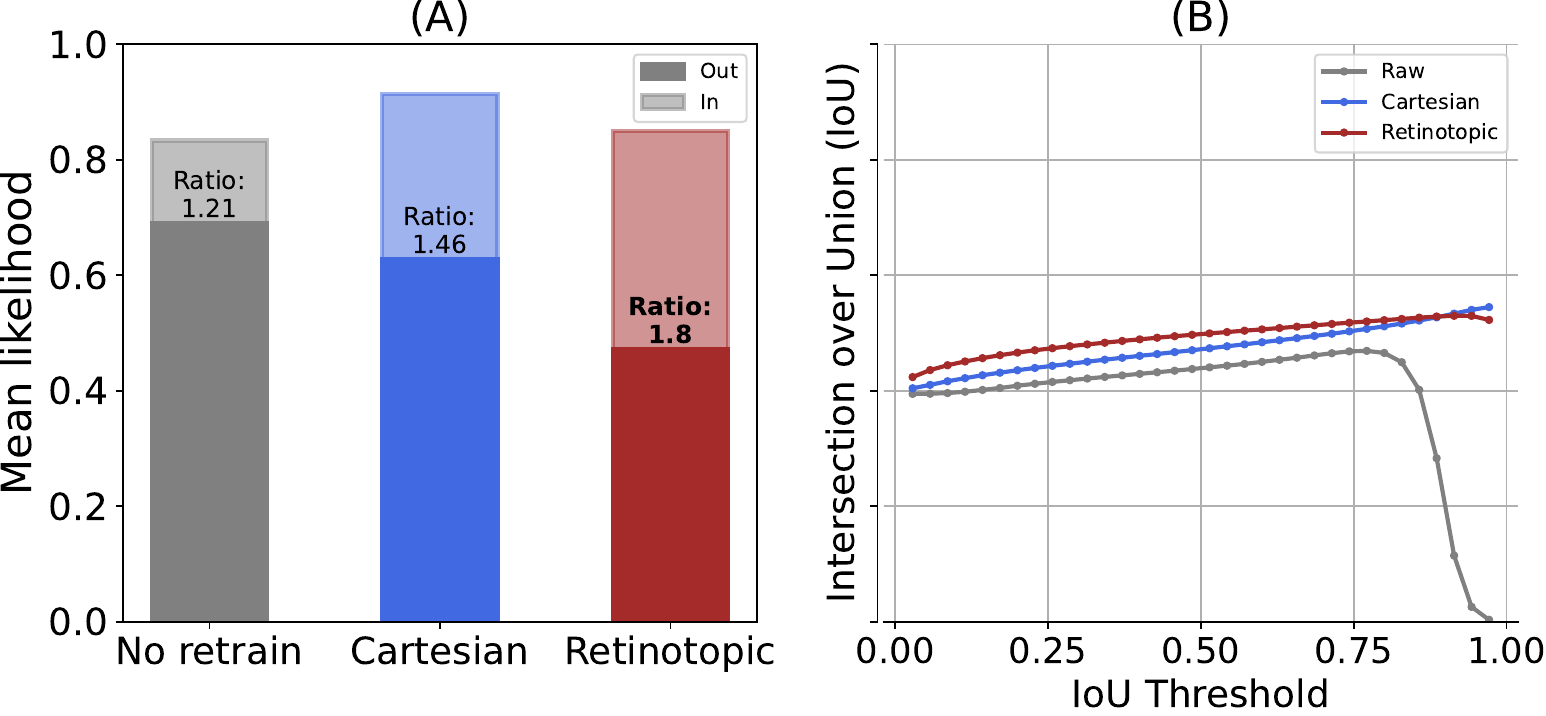}
         \caption{We tested the visual search protocol on the {\sc Animal 10k}  (comprising $10,015$ images). {\bf (A)}~The mean likelihood across the point of fixation inside the bounding box (``In'') or the point of fixation outside the bounding box (``Out'') and the corresponding likelihood ratio. {\bf (B)}~The intersection over union as a function of a threshold applied on the likelihood map. }    
         \label{animal_10k_mesure}
   \end{center}
         
\end{figure}

\subsection*{Beyond {\sc ImageNet}, the {\sc Animal-10k} dataset} 
{\sc ImageNet-1K} provides rich semantic links that allow the construction of task-specific datasets. It has previously been demonstrated that the use of fine-tuning to re-train networks such as {\sc Vgg16}~\citep{Simonyan2015} allows them to be applied to different tasks using the semantic network underlying {\sc ImageNet}'s labels. Furthermore, it has been shown that the probability of a trained network performing a novel task (such as categorizing an animal) can be predicted using this semantic network, which links the outputs of a pre-trained network to a library of labels. This approach has been shown to be an effective method for learning to predict the presence of animals in images~\citep{Jeremie2023}. We can then exploit the semantic connection underlying the {\sc ImageNet} labels by using these networks to perform a categorization task on the second dataset, the {\sc Animal 10k} dataset, as referenced by~\citet{Yu2021}, with the goal of identifying the animals depicted in the images. This dataset contains $10,015$ images, each depicting an animal. All images were integrated in a validation dataset to test the networks. The output of each network is interpreted as ``animal'' if the label represented by the highest likelihood after softmax is indeed an animal. We use this prediction to calculate the accuracy of predicting an animal for all networks. Note that the accuracy is slightly biased due to the absence of distractors, it only represents the probability of finding an animal when there is actually an animal in the input. Figure~\ref{pointing_game_accuracy} (right panel) compares the accuracy improvement for both networks. The general accuracy is high (around $95\%$ in all cases) due to our simpler binary (animal/non-animal) classification task. Saccade improvement is still significant, with all networks reaching a maximal accuracy of $100\%$ after a saccade to the maximum likelihood position for the label ``animal''. The highest improvement is once again found in the retinotopic case (``raw'' $4.5\%$, Cartesian $4.1\%$, retinotopic $6.3\%$).

Regarding the localization capability, the In/Out likelihood ratio (Figure~\ref{animal_10k_mesure}A) shows clear contrast enhancement, from 1.2 (raw network) toward 1.5 (Cartesian ``focus'' network) and 1.8 (Retinotopic ``focus'' network). The general lower contrast, when compared to {\sc ImageNet}, comes again from the simpler binary (animal/non-animal) classification task, making it possible to guess the label ``Animal'' outside the bounding box from the background information.
Looking now at the IoU metric (Figure~\ref{animal_10k_mesure}B), a different trend is observed, with monotonically increasing IoU with the likelihood map threshold, reflecting a convergent matching of the likelihood map with the animal contour. The optimal threshold, close to 1, reflects a clear and sharp transition in both cases. Here, the retinotopic network provides slightly higher IoU values, except at the highest threshold case where the Cartesian network prevails. The difference between this and {\sc ImageNet} lies in the fact that masks are used in the calculation instead of boxes (see Methods).
Last, the pointing game results (see Figure~\ref{pointing_game_accuracy}B, green bars) provide a clear assessment of the Cartesian/retinotopic contrast. While both the raw and Cartesian network remain quite low (73-74\% success), the retinotopic framework shows much higher pointing success (87.7\%) despite refined animal contour, once again suggesting a more reliable animal localization. Despite the increased diversity of poses, environments, and intra-class variability in {\sc Animal 10k} compared to {\sc ImageNet-1K}, our results indicate that the regions containing the target animals are effectively identified.

\begin{figure}[!t]
  \includegraphics[width=1\linewidth]{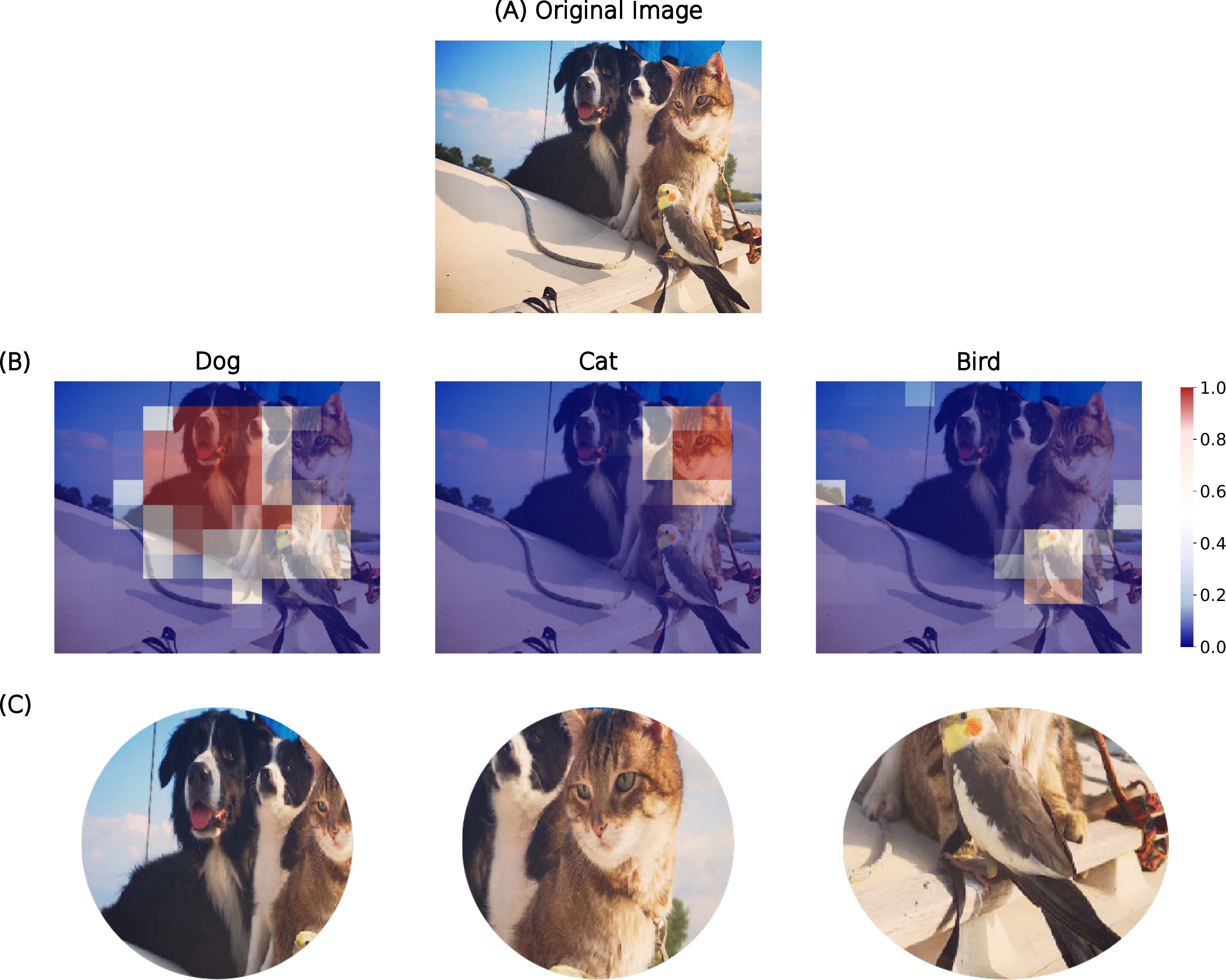}
  \caption{The likelihood maps presented here were generated by {\sc Resnet101} networks that were trained and tested on retinotopic space using the log-polar grid. {\bf (A)}~The original image, it is processed on a single iteration, with the likelihood maps displayed in accordance with the specific combinations of labels of interest. {\bf (B)}~Likelihood maps indicating the probability of the presence of the labels ``dog'', ``cat'' and ``bird'' in the image. {\bf (C)}~The sample derived from all fixation points that yield the highest likelihood value for a label of interest.}
  \label{multi_label}
\end{figure}

 \subsubsection*{Multi-label and multi-task extension}
The likelihood map can be extended along the dimension of labels. The networks perform a discriminative categorization among $1000$ {\sc ImageNet} labels, and the likelihood maps generated by the model vary depending on the label or class being predicted for a given input image. Specifically, the spatial distribution of high likelihood regions differs according to the object category the model is trying to identify. This suggests that the model learns to focus on different discriminative regions for different classes, providing insight into how the model's spatial attention adapts to the visual patterns that drive each classification decision. We tested this hypothesis on an example image by generating likelihood maps for the classes ``dog'', ``cat'', and ``bird''. As shown in Figure~\ref{multi_label}, the peak activations in these maps appear in distinct spatial regions corresponding to the locations of each respective animal in the image.

\section*{Discussion}

 \subsection*{Retina-inspired mapping enhances the robustness of CNNs}
The first and main result of this study is to demonstrate the excellent ability of standard deep CNNs to deal with foveated retinotopic inputs, even though this mapping enforces a radical transformation of the visual inputs. {\sc ResNet} networks easily adapt to these inputs and the accuracy rates achieved with retinotopic inputs are equivalent to those of the original models. This is surprising given that the networks used in this re-training process were previously trained on Cartesian images. Images with a log-polar transformation show a high degree of distortion, particularly a high compression of visual information around the fixation point and a degradation of textures in the periphery (see Figure~\ref{logpolar}). One possible hypothesis is that the degradation of texture during the frame of reference change may cause the network to rely on shape rather than texture~\citep{da_costa_convolutional_2024}. A further study could entail a comparison of these visual processing parameters with those of human vision, thereby providing insight into their evolutionary trajectory.

In addition, the log-polar transformation has the advantage of better invariance to zooms and rotations. First, this study shows that for the original {\sc ResNet} network on ``regular'' images, the average accuracy with a zoom or a rotation dropped sharply compared to baseline accuracy, confirming that these simple geometrical transformations misled the networks. For rotations, the decline was steepest around $160^\circ$, demonstrating limited rotational invariance compared to humans~\citep{Rousselet2003,Guyonneau2006}. The robustness differs slightly when studying the impact of zooms. This may be attributed to the transformation under examination, which already provides a zooming-in effect of the fovea that accentuates the information in the area surrounding the point of fixation. To test this hypothesis, we could train a retinotopic version without applying the logarithmic function to the eccentricity axis. Finally, this invariance to rotations and zooms comes at the cost of reduced invariance to translations. For images not centered on the region of interest, one would need to shift the fixation point to the region of interest, similar to eye saccades in biological vision. The effects of this trade-off can be observed in the comparison between the ``regular'' version of the networks and the ``focus'' version.

In addition, the log–polar transformation confers improved invariance to zoom and rotation. In the original {\sc ResNet} trained on ``regular'' Cartesian images, accuracy declined sharply under zooms or rotations, confirming that even simple geometric transformations can mislead the network. The steepest drop occurred around $160^\circ$, highlighting limited rotational invariance compared to humans~\citep{Rousselet2003,Guyonneau2006}. Robustness to zoom was less affected, likely because the transformation itself magnifies foveal input relative to the periphery. This robustness, however, comes at the cost of reduced tolerance to translations: images misaligned with the fixation point require shifting the sampling center, analogous to saccades in biological vision. This trade-off is evident when comparing the ``regular'' and ``focus'' network variants. Similar effects have been reported previously: ~\citet{pun_log-polar_2003} showed that log–polar wavelet signatures enhance joint scale–rotation invariance, while ~\citet{amorim_analysing_2018} demonstrated improved rotational robustness in natural image categorization and highlighted the positional sensitivity introduced by inhomogeneous log–polar sampling. Finally, previous research has demonstrated that improved invariance to scale variations is often associated with greater robustness to input compression~\citep{Remmelzwaal2019}. This dual robustness — the ability to maintain performance across scaled inputs and compressed representations — facilitates the application of categorisation models to high-resolution images without compromising accuracy. These properties are particularly valuable in computational neuroscience, where visual processing systems must generalise across diverse input conditions.
%

 \subsection*{From Foveation to Pre-attentive Mechanisms}
The second result of this study is the emergence of localization properties in networks re-trained with foveated inputs. Through the definition of likelihood maps, which collate the model's output while scanning the visual scene at a limited number of fixation points, we gain insight into the specificities of retinotopic processing. This transformation provides a more focused view, thus better separating different elements of the image when focusing on specific parts. Such processing is reminiscent of pre-attentive mechanisms which allow biological vision to selectively process important zones of the visual space. As a result, retinotopic transformation provides a proxy for the measurement of saliency, particularly with respect to a set of cued labels.

One hypothesis is that the foveated retinotopic mapping implemented in the log-polar transform provides an efficient prior for visual object geometry. Indeed, a main source of variability in the view of an object comes from displacements of the observer relative to it, for example as the object scales in the visual field as one approaches it, or as the object rotates with a rotation of the head. The log-polar mapping allows for a more invariant representation which explicitly implements this prior for these displacements. Additionally, imposing translation invariance of the representation establishes a prior on the possible representations allowed by the network~\citep{Lempitsky2018}. However, it has been observed that different classes have different statistics, and the relative size of buses is on average larger than that of cats. Consequently, we expect that different foveated retinotopies may emerge in different ecological niches.

In our case, it seems that the foveated retinotopic mapping allows for a more precise localization of the category of interest compared to off-the-shelf pre-trained networks using a Cartesian representation. It also gives us insight into the features on which our networks actually rely. Such information can be compared with physiological data~\citep{Crouzet2011}, used to design better CNNs, and ultimately enable the development of physiological tests to further explore the features needed to classify a label of interest. In particular, by focusing on the point of fixation with the highest probability in the likelihood maps instead of the center of the bounding box defined in the ``focus'' dataset, we could consider refining the training of the network using our retinotopic mapping in a semi-supervised fashion.

 \subsection*{Perspectives and future work}
Building on these observations, simulating human-like saccadic eye movements offers a powerful approach to probe network mechanisms. A protocol that iteratively classifies foveated image patches could approximate natural viewing and reveal how performance varies across the visual field, as hypothesized in psychophysical experiments~\citep{han_scale_2020}. Comparing accuracy under different saccadic planning strategies—for example, prioritizing the most uncertain versus the most likely locations—would shed light on network attentional dynamics. This framework also enables modeling of biologically inspired saccade generation, allowing direct comparison with human visual search behavior.

Overall, implementing foveated classification with algorithmic saccades would provide a powerful method for validating existing attentional mechanisms in these networks, as well as inspiring new architectural innovations through embodied, task-driven visual attention modeling. In particular, this line of research would be especially suitable for dual-pathway models that effectively infer ego-motions~\citep{Bakhtiari2021,Mineault2021}. Finally, the implementation of this robust categorization, coupled with refined localization of a label of interest and optimal saccade selection, could allow us to extend this study to more complex tasks. Visual search (i.e., the simultaneous localization and detection of a visual target) represents one such application where likelihood maps could provide the underlying pre-attentive mechanisms on which its effectiveness depends.
\section*{Methods}
\subsection*{The log-polar transform}
In most mammals and amphibians, the arrangement of the (external) visual field is preserved in the early visual pathway, a feature called retinotopy. Retinotopic mapping results from the combined effect of the arrangement of photoreceptors in the retina and their output convergence via the optic nerve. This causes nearby regions of the visual field to activate adjacent neural structures as signals travel from the retina to the brain. These mappings differ from species to species, and our study concentrates on foveated vision (as in humans) which gives more resolution to the central field of view. In particular, we implement it by transforming the Cartesian pixel coordinates into log-polar coordinates~\citep{JavierTraver2010}.

This simple parameterized transformation models accurately this biologically inspired retinotopic mapping. Considering arbitrary images (potentially with multiple channels such as RGB), each pixel's position is defined by two real coordinates $(x,y)$ on a Cartesian reference frame. By convention, $x$ and $y$ are here considered belonging to the interval $[-1,1]$, with $(0,0)$ being the center of the image. To implement the concentration of pixels near the center of the retina, we need to consider an \emph{irregular} grid in the Cartesian referential that maps to a \emph{regular} grid in the log-polar referential. In the log-polar referential, the location of each pixel has corresponding coordinates (${\rho}, {\theta}$) as defined by previous studies~\citep{Hao2021} by:
\begin{eqnarray}
   \rho &=& \log_2 \sqrt{(x-x_0)^2+(y-y_0)^2} \\
   \theta &=& \arctan( \frac{y-y_0}{x-x_0} )
\end{eqnarray}
with $(x_0,y_0)$ defining the ``center of fixation'' (see Figure~\ref{logpolar})~\citep{Traver2003}.  Importantly, $\rho$ and $\theta$ are only defined for $(x,y) \neq (x_0,y_0)$.  In most of our experiments, we consider $x_0=(0,0)$, allowing us to focus on the central part of the image. Each dot ($\rho$, $\theta$) in the log-polar coordinate system has thus a unique correspondence in the Cartesian coordinate system (and vice versa): For each ($\log \rho$, $\theta$) belonging to the grid, the corresponding pixel coordinate is $(x_0 + \rho \cos \theta, y_0 + \rho \sin \theta)$. 

In practice, images have a finite resolution and to avoid biases in the evaluation between networks, the number of angles sampled ($N_\theta$) and the number of eccentricities sampled ($N_\rho$) are set to $224$, so that the size of the transformed image match the resolution $224\times224$ of the input images. Note that native image resolution is generally higher than that used during network processing, with the average resolution in the Imagenet dataset being around $500\times500$ pixels. This transformation is performed with the {\sc PyTorch} library~\citep{Paszke2019} through the use of the \lstinline{grid_sample()} function, which maps the pixels of an input image to the coordinates of any arbitrary grid, using a linear interpolation to estimate the value of the pixels. This function is used, for instance, in spatial transformer networks~\citep{Jaderberg2016}. 

Let's now define each coordinates. All $\theta$ values are within a linear distribution in $[0;2\pi)$, while $\rho$ values are within a logarithmic interval with $r_\text{min}$  the minimal radius and $r_\text{max}$ be the maximal radius (with $N_\rho$  the radial resolution). In practice, we use a log-polar grid with an outer log-radius of $\log_2 r_\text{max} = 0$ ($r_\text{max}=1$, defining a circle tangent to the image box) and an inner log-radius of $\log_2 r_\text{min} = -5$ ($r_\text{min}=2^{-5}$).  In summary, the regular grid is the interval $[\log_2 r_\text{min}, ..., \log_2 (r_\text{min} + i\times \frac{r_\text{max}-r_\text{min}}{N_\rho-1}) ,...,\log_2 r_\text{max}]$, for $i$ in $[0,..., N_\rho-1]$, in the $\log_2 \rho$ dimension and $[0, ..., j\times\frac{2\pi}{N_\theta} ,..., 2\pi (1 - \frac{1}{N_\theta})]$, for $j$ in $[0,..., N_\theta-1]$ in the $\theta$ dimension. Note that some pixels in the log-polar grid may be smaller than the pixels from the Cartesian grid.

\subsection*{Convolutional Neuronal Networks (CNNs)}

Convolutional neural networks (CNNs) have become essential tools in image classification, with several pre-trained models available for download. For example, the VGG family, including VGG16 and VGG19, introduced by~\citet{Simonyan2015}, uses deep architectures with $16$ or $19$ layers, consisting of stacked convolutional layers followed by fully connected layers. {\sc ResNet} (Residual Networks), introduced by~\citet{He2015wrn}, addresses the vanishing gradient problem in deep networks by incorporating skip connections, allowing for the training of extremely deep networks (e.g. {\sc ResNet} $50$, {\sc ResNet} $101$ with respectively $50$ and $101$ layers). These models are widely used due to their robustness and scalability. The implementation of these deeper networks have demonstrated that deeper networks display enhanced resilience; however, this improvement is coupled with an overall increase in computational complexity~\citep{He2015wrn}.  Therefore, based on these findings, we focus on the deep CNN {\sc ResNet} (with $18$ to $101$ layers from the {\sc PyTorch} library on the {\sc ImageNet-1K}~\citep{Russakovsky2015} categorization challenge which consists in classifying natural images into $1000$ labels. We further introduced a circular padding in the convolutions, however we controlled this had little impact overall.

\subsection*{Datasets}

\begin{figure}[ht!] 
   \begin{center}
      \includegraphics[width=\textwidth]{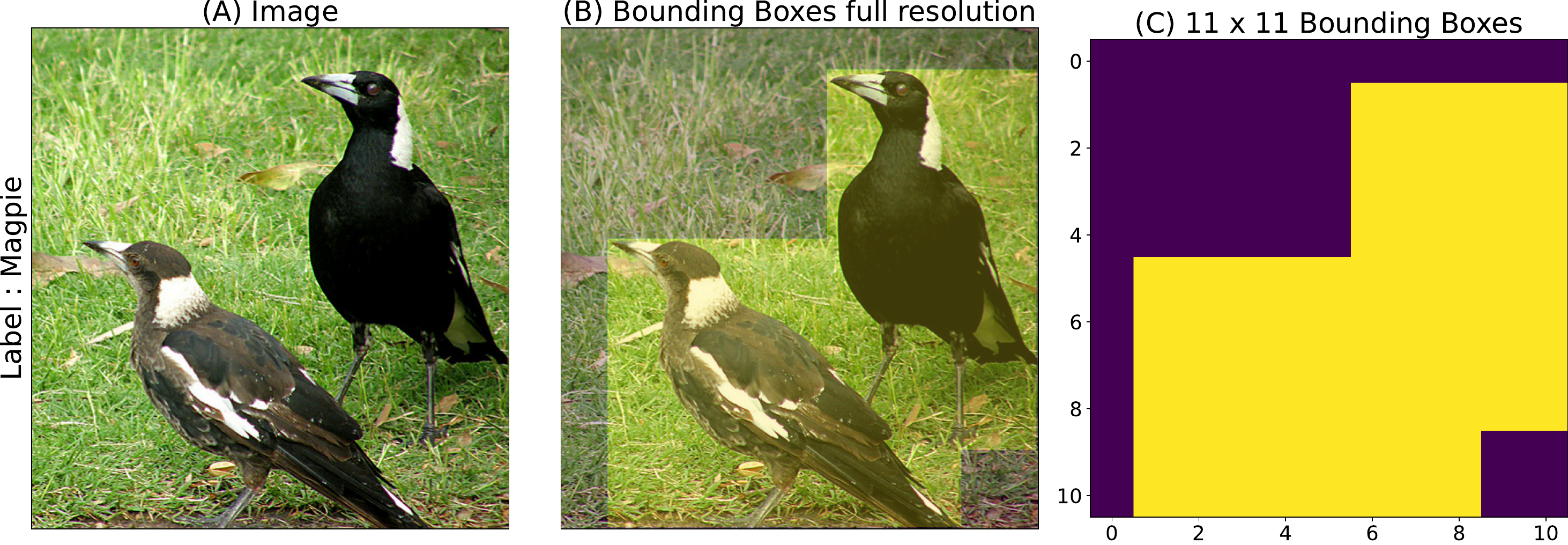}
   \end{center}
   \caption{
      {\bf (A)}~Original image from the ImageNet dataset. {\bf (B)}~Bounding-box annotation shown as a heat map. {\bf (C)}~Downsampled heat map ($11 \times 11$) used as ground truth for localization. The ``regular'' dataset uses the full image as input (Cartesian or log-polar for the retinotopic framework), whereas the ``focus'' dataset uses the cropped bounding box of the target label (yellow boxes in B). To ensure consistency, if there are multiple boxes present, only the first annotation provided in the dataset is used when constructing the ``focus'' dataset.
      } 
      \label{imagenet_ground_truth}
   \end{figure}
Typical image classification datasets used in machine learning consist of sets of RGB images of different resolutions, each image being associated with a single label. The classification task involves learning a parametric function that learns to associate a unique label with a set of (high-dimensional) pixels. Example-based learning therefore assumes that there are regions within the representation space (or feature space) that can be learned in order to separate objects into these different classes, regardless of their position in the image, size, orientation, lighting, contrast etc...

Two datasets were used for our study: the first dataset is the one from the {\sc ImageNet}~\citep{Russakovsky2015} challenge, which is the most widely benchmarked due to its huge collection of images and associated labels (the subset of {\sc ImageNet} used in this study, i.e. {\sc ImageNet-1K} with $1000$ labels and about $1000$ examples per label). In addition to the classification task, we consider here the localization task, which is prominent in computer vision. It consists in identifying both the label and the position of one or more objects of interest in an image. A distinction can be made between (i) the \emph{visual search} task, where the label is given in advance and the task is simply to find the position of the object in the image, and (ii) the \emph{image labeling} task, which consists in identifying both the objects and their position in the image, in the form of a labeled bounding box (see Figure~\ref{imagenet_ground_truth}). Despite its advantages, {\sc ImageNet} has limitations for localization tasks. For example, the dataset lacks multi-labeling, with only one label per image. It is worth mentioning that {\sc ImageNet} has some biases, the objects of interest are generally centered in the images, and the proportion of bounding boxes relative to the image size is relatively large, which may limit the impact of certain analyses. 

We consider two dataset configurations for the {\sc ImageNet} dataset: In our first configuration, the center of gaze is set to the center of the original image, taking advantage of the fact that most {\sc ImageNet} images are human-made and that photographers have a bias toward centering the object of interest. This defines our ``regular'' image dataset. Notwithstanding the above, this a priori assumption of centered position is not sufficient to generate a dataset perfectly suited for retinotopic transformation. In a second setup (called the ``focus'' dataset), we use the bounding box information provided by the {\sc ImageNet} dataset. Consequently, a sample is selected, defined as the smallest square containing a bounding box, assuming a center of gaze now at the center of the bounding box for the label of interest. This novel dataset is used to train a second generation of networks. Again, we use a circular mask for the Cartesian frame. This approach is more robust to the position of the visual object, but requires reliable bounding boxes.

\begin{figure}[ht!] 
   \begin{center}
      \includegraphics[width=\textwidth]{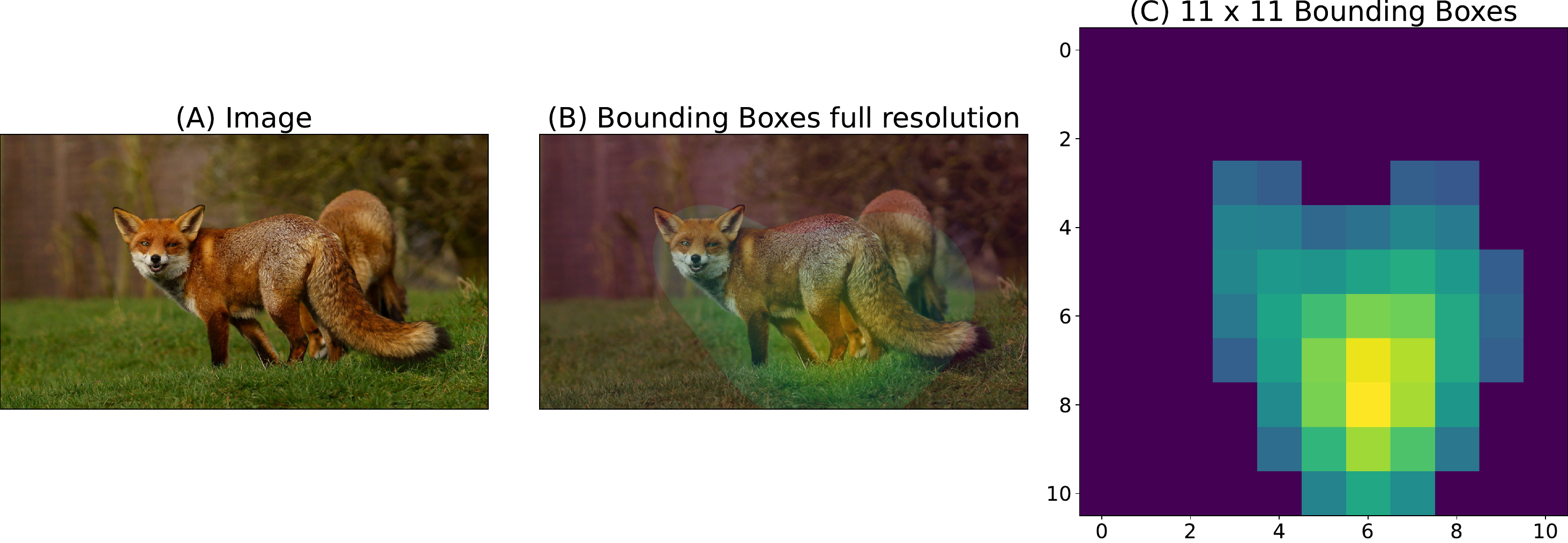}
   \end{center}
   \caption{
      {\bf (A)}~Original image of the {\sc Animal 10k} dataset. {\bf (B)}~Heat map generated by fitting Gaussians to annotated key points. {\bf (C)}~Heat map from {\bf (B)}~normalized and downsampled to $11 \times 11$, used as ground truth. A threshold of 0.2 is applied to delineate the assumed animal contour.
      } 
      \label{animal_10k_ground_truth}
   \end{figure}

The dataset provides a set of key points for each animal present in an image. For each image from {\sc Animal 10k} containing a set of keypoints, we created a Gaussian heatmap centered on those points, with the peak value set to 1 and values decreasing with a standard deviation proportional to object size, thus capturing the true spatial extent and location of the target animal within each image (see Figure~\ref{animal_10k_ground_truth}). This approach allows for more effective localization and analysis of the visual distribution of animals in images.

\subsection*{Datasets transformations and Transfer learning}

In this study, three series of transformations were used, depending on whether the network used the Cartesian or the retinotopic reference frame. In the case of the latter, a log-polar grid is used to facilitate the transformation of the image into the retinotopic frame. Due to the intrinsic nature of the transformation, which results in the cropping of a circular sample within the original image, a circular crop is implemented for the Cartesian frame to ensure comparability. To allow for a more straightforward comparison, the ``raw'' datasets were processed without a circular mask or polar logarithmic transformation. Each set underwent a uniform transformation, including normalization to tensors and resizing to a resolution of $244\times244$ to match the pre-trained parameters of the model. 

To assess the efficacy of our retinotopic mapping, we examine popular off-the-shelf CNNs pre-trained on standard, large image datasets. These networks are re-trained on our datasets, either with (or without) a log-polar retinotopic transformation, using the cross-entropy loss from the {\sc PyTorch} library. We use the stochastic gradient descent (SGD) optimizer from the {\sc PyTorch} library and validate parameters such as batch size, learning rate, and momentum by performing a sweep of these parameters for each network. During the sweep, we vary each of these parameters over a given range while leaving the others at their default values for $1$ epoch on 10\% of the entire {\sc ImageNet} training dataset. We choose the parameter values that give the best average accuracy on the validation set: batch size = $80$, learning rate = $0.001$, momentum = $0.9$. We re-trained the networks during $2$ epochs of the full training dataset, keeping all learning parameters identical.

\subsection*{Attacking classical CNNs with a geometrical rotation}

A common approach to evaluating the robustness of deep learning models is to subject them to adversarial attacks. In this study, we investigate the robustness of the deep learning models to natural image transformations that are easily perceived by humans. In particular, we evaluated the performance of the networks on the {\sc ImageNet} dataset when the images were rotated by different angles and averaged the accuracy for each angle. To further evaluate the robustness to rotations, we also designed a ``rotation-based attack'' scenario. To perform such an attack on a model $m$, we follow this simple procedure. Given an image $I$ and the output of the model $\vp = m(I)$, which returns a probability vector over $K=1000$ classes, the loss function $\mathcal{L}$ is defined as the cross-entropy between the predicted probability vector and the ground truth label $y$, which we denote as $\mathcal{L}(m(I), y)$. This is the loss minimized during gradient descent training. We then systematically rotated the images and tracked the change in model loss. By denoting a rotation of the image by an angle $\theta$ as $\text{rot}(I; \theta)$, we define the rotation-based attack as the following heuristic for each image in the dataset: 
\begin{eqnarray}
   \bar{\theta} &= &\argmax_{\theta} \mathcal{L}(m(\text{rot}(I; \theta)),y) \\
   \hat{y} &= & \argmax_k ( \bar{\evp}_k ) \text{ with } \bar{\vp} = m( \text{rot}(I; \bar{\theta}) )
\end{eqnarray}
More specifically, our approach is to first choose the rotation angle that maximizes the loss, and then infer the most likely label for that particular angle. As a result, we can compute the concordance between the predicted label $\hat{y}$ for the image rotated at the angle $\bar{\theta}$ with the worst loss with respect to the ground truth label $y$. Using this procedure, we calculated the overall accuracy on the entire test set, quantifying the network's brittleness to natural image rotations. We use a similar strategy for other geometric transformations, such as zooms or translations. 

\subsection*{Localization tools and evaluation}

A widely accepted technique for evaluating the performance of Convolutional Neural Networks (CNNs) in localization tasks is the Class Activation Mapping (CAM) method. CAM works by analyzing the output of the CNN with respect to the target class, assigning weights to activations in each spatial feature map. This process generates a heat map that highlights significant areas of the image based on their contribution to the prediction. Building on the foundation of CAM, several derivative methods have emerged, including Grad-CAM~\citep{Selvaraju2020}, Score-CAM~\citep{Wang2020}, and Opti-CAM~\citep{Zhang2024}.

In an effort to fairly quantify the respective contributions of each method, many quantification techniques have been developed. Here we select some of them to compare the models using the retinotopic or Cartesian reference. {\sc Energy-Based Pointing Game}: Localization is successful if the peak activation of the heatmap of a given label is inside the ground true mask (or box). {\sc Mean Activation IN}: Mean activation of the heatmap of a given label inside the ground true mask. {\sc Mean Activation OUT}: Mean activation of the heatmap of a given label outside the Ground True mask. {\sc Mean Activation Ratio}: Ratio of activation inside and outside the box; the higher the value, the more efficient the heatmap is at indicating the position for a given label. {\sc Intersection over Union (IoU)}: Ratio of the area of overlap between the heatmap and the ground truth to the area of union between the heatmap and the ground truth. {Peak-IoU and Peak-Threshold}: For a modulation of a threshold on the heat map, the Peak-IoU is the maximum IoU value reached at the Peak-Threshold.

\begin{figure}[ht]
\begin{center}
\includegraphics[width=1\textwidth]{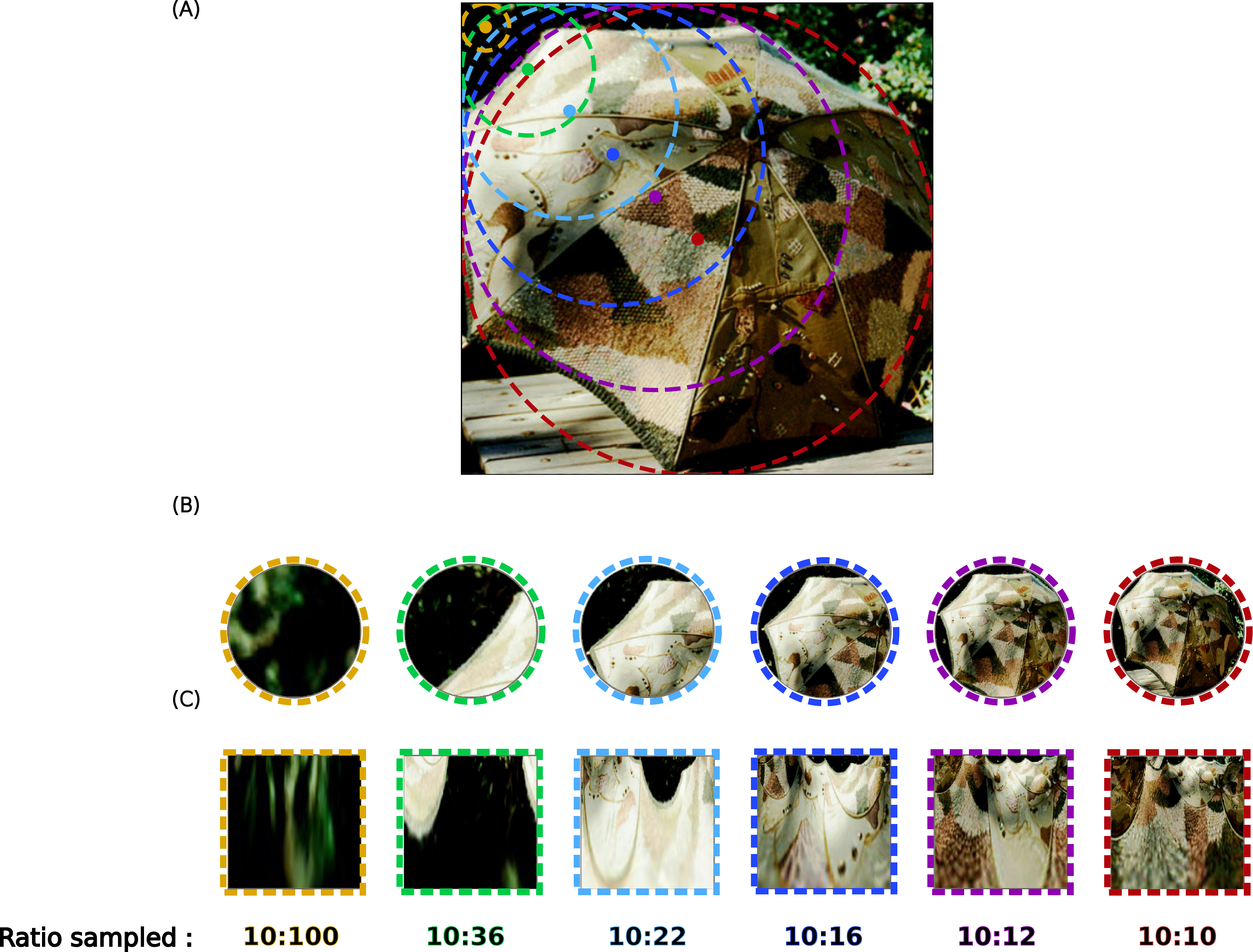}
\end{center}
\caption{We illustrate the protocol used to obtain the likelihood map with an example on a single image {\bf (B)}. We show one sample for each sample ratio size ($1:10$ to $1:1$, from left to right). {\bf (A)}~The samples are cropped on the image in the Cartesian frame, we use circular cropping to match the area covered by the log-polar frame. {\bf (C)}~The corresponding samples cropped on the image in the retinotopic frame.} 
\label{likelihood_map_protocol}
\end{figure}

\subsection*{Visual object localization : Protocol}

Deep convolutional networks such as {\sc ResNet} output a vector of real numbers which predict the logit of the likelihood in label space, and this prediction is optimized through cross-entropy loss. Applying the softmax function allows the output to be interpreted as a probability vector, assigning a probability of presence to each of the $1000$ labels (likelihood score). 

This allows us to make a binary decision (``present'' or ``not present''), e.g. by selecting the label corresponding to the highest likelihood. In our setting, we can also take different views of a large image and compute the likelihood for each of them, allowing us to compare which view provides the best likelihood. Views can consist, for example, of focusing on regions of the image centered on different fixation points, with the fixation points aligned on a regular grid in visual space.

We used two parameters to define these maps. The first parameter is the resolution of the fixation point grid, and when not specified this resolution is set to $11\times11$. The second is the size of the samples clipped at each of these positions, defined as the ratio of the log-polar grid radius of the input to the total input size or Cartesian grid size, since the grid is a square for Cartesian samples (for an illustration of this simple process, see Appendix Figure~\ref{likelihood_map_protocol}). At each viewpoint, the largest possible sample is cropped. Thus a minimum sample with $1:10$ ratio at the border and the whole image at the center. From the Cartesian or retinotopic reference frame, this sample is then resized, if necessary, to a $224\times224$ resolution to match the input size for the CNN before processing or transformed by the retinotopic mapping (also at $224\times224$ resolution) before being used as input for the corresponding network, see Appendix Figure~\ref{likelihood_map_protocol}-B \& D. Conveniently, a collection of samples for different fixation points can be processed as a single batch. This protocol defines a likelihood map for any given network as the likelihood of categorizing the presence of a label of interest inferred at regularly spaced fixation points in the image. 

\section*{Declarations}

\subsection*{Data Availability}
All results presented here are based on data that are openly available, as described in the Methods section.

\subsection*{Code Availability}
The code to reproduce all results and figures is available at
\url{https://github.com/bicv/Retinotopy}.

\subsection*{Acknowledgments}

Authors received funding from the ANR project number ANR-20-CE23-0021 (\enquote{\href{https://laurentperrinet.github.io/grant/anr-anr/}{AgileNeuRobot}}) and from the french government under the France 2030 investment plan, as part of the Initiative d’Excellence d’Aix-Marseille Université – A*MIDEX grant number AMX-21-RID-025 \enquote{\href{https://laurentperrinet.github.io/grant/polychronies/}{Polychronies}}.

This work was granted access to the HPC resources of Aix-Marseille Université financed by the project Equip\verb|@|Meso (ANR-10-EQPX-29-01) of the program \enquote{Investissements d’Avenir} supervised by the Agence Nationale de la Recherche.

For the purpose of open access, the authors have applied a CC‑BY public copyright licence to any Author‑Accepted Manuscript version arising from this submission.

\subsection*{Author Contributions}

All authors contributed equally to the following roles according to the CRediT Taxonomy: Conceptualization, Methodology, Software, Validation, Formal Analysis, Investigation, Resources, Data Curation, Writing - Original Draft, Writing - Review \& Editing, and Visualization. LP and ED provided Supervision and LP performed Project Administration, and Funding Acquisition. All authors have read and agreed to the published version of the manuscript.

\subsection*{Competing Interests}

We affirm that this manuscript is original, has not been published, and is not under consideration elsewhere. All authors have read and approved the final version of the manuscript. We have adhered to ethical research guidelines and declare no conflicts of interest.

\bibliographystyle{plainnat}

\bibliography{jeremie_2025_retinotopy}

@misc{Bengio2016,
  title = {Towards {{Biologically Plausible Deep Learning}}},
  author = {Bengio, Yoshua and Lee, Dong-Hyun and Bornschein, Jorg and Mesnard, Thomas and Lin, Zhouhan},
  year = {2016},
  month = aug,
  number = {arXiv:1502.04156},
  eprint = {1502.04156},
  primaryclass = {cs},
  publisher = {arXiv},
  doi = {10.48550/arXiv.1502.04156},
  urldate = {2024-12-29},
  archiveprefix = {arXiv}
}

@misc{Mineault2024,
  title = {{{NeuroAI}} for {{AI Safety}}},
  author = {Mineault, Patrick and Zanichelli, Niccol{\`o} and Peng, Joanne Zichen and Arkhipov, Anton and Bingham, Eli and {Jara-Ettinger}, Julian and Mackevicius, Emily and Marblestone, Adam and Mattar, Marcelo and Payne, Andrew and Sanborn, Sophia and Schroeder, Karen and Tavares, Zenna and Tolias, Andreas},
  year = {2024},
  month = nov,
  number = {arXiv:2411.18526},
  eprint = {2411.18526},
  primaryclass = {cs},
  publisher = {arXiv},
  doi = {10.48550/arXiv.2411.18526},
  urldate = {2024-12-03},
  archiveprefix = {arXiv}
}

@misc{Bashivan2023,
  title = {Learning {{Robust Kernel Ensembles}} with {{Kernel Average Pooling}}},
  author = {Bashivan, Pouya and Ibrahim, Adam and Dehghani, Amirozhan and Ren, Yifei},
  year = {2023},
  month = may,
  number = {arXiv:2210.00062},
  eprint = {2210.00062},
  primaryclass = {cs},
  publisher = {arXiv},
  doi = {10.48550/arXiv.2210.00062},
  urldate = {2024-09-11},
  archiveprefix = {arXiv}
}

@article{Martins2024,
  title = {Building egocentric models of local space from retinal input},
  volume = {34},
  issn = {0960-9822},
  url = {https://www.cell.com/current-biology/abstract/S0960-9822(24)01452-0},
  doi = {10.1016/j.cub.2024.10.057},
  language = {English},
  number = {23},
  urldate = {2024-12-10},
  journal = {Current Biology},
  author = {Martins, Dylan M. and Manda, Joy M. and Goard, Michael J. and Parker, Philip R. L.},
  month = dec,
  year = {2024},
  pmid = {39626632},
  note = {Publisher: Elsevier},
  pages = {R1185--R1202},
}

@article{Faramarzi2024,
  title = {A 128×128 {Electronically} {Multi}-{Foveated} {Dynamic} {Vision} {Sensor} with {Real}-{Time} {Resolution} {Reconfiguration}},
  issn = {2169-3536},
  url = {https://ieeexplore.ieee.org/document/10804122/?arnumber=10804122},
  doi = {10.1109/ACCESS.2024.3519035},
  journal = {IEEE Access},
  author = {Faramarzi, Farnaz and Linares-Barranco, Bernabé and Serrano-Gotarredona, Teresa},
  year = {2024},
  note = {Conference Name: IEEE Access},
  pages = {1--1},
}

@unpublished{Lu2023,
  title = {End-to-End Topographic Networks as Models of Cortical Map Formation and Human Visual Behaviour: Moving beyond Convolutions},
  shorttitle = {End-to-End Topographic Networks as Models of Cortical Map Formation and Human Visual Behaviour},
  author = {Lu, Zejin and Doerig, Adrien and Bosch, Victoria and Krahmer, Bas and Kaiser, Daniel and Cichy, Radoslaw M. and Kietzmann, Tim C.},
  year = {2023},
  month = aug,
  eprint = {2308.09431},
  eprinttype = {arXiv},
  eprintclass = {q-bio},
  doi = {10.48550/arXiv.2308.09431},
  url = {http://arxiv.org/abs/2308.09431},
  urldate = {2024-12-20},
  archiveprefix = {arXiv}
}

@article{Traver2003,
  title = {Designing the {Lattice} for {Log-Polar Images}},
  author = {Traver, V Javier and Pla, Filiberto},
  year = {2003},
  journal = {Discrete Geometry for Computer Imagery},
  pages = {164--173},
  issn = {03029743},
  url = {http://link.springer.com/10.1007/978-3-540-39966-7_15}
}

@inproceedings{Sarvaiya2009,
  ids = {Sarvaiya2009a},
  title = {Image Registration Using Log-Polar Transform and Phase Correlation},
  booktitle = {{{IEEE Region}} 10 {{Annual International Conference}}, {{Proceedings}}/{{TENCON}}},
  author = {Sarvaiya, Jignesh N. and Patnaik, Suprava and Bombaywala, Salman},
  year = {2009},
  pages = {1--5},
  doi = {10.1109/TENCON.2009.5396234}
}

@article{Araujo1997,
  title = {An Introduction to the Log-Polar Mapping},
  author = {Araujo, H. and Dias, J.M.},
  year = {1997},
  journal = {Proceedings II Workshop on Cybernetic Vision},
  number = {1},
  pages = {139--144},
  doi = {10.1109/CYBVIS.1996.629454},
  url = {http://ieeexplore.ieee.org/document/629454/}
}

@article{da_costa_convolutional_2024,
  title = {Convolutional neural networks develop major organizational principles of early visual cortex when enhanced with retinal sampling},
  volume = {14},
  rightsholder = {The Author(s)},
  issn = {2045-2322},
  url = {https://www.nature.com/articles/s41598-024-59376-x},
  doi = {10.1038/s41598-024-59376-x},
  pages = {8980},
  number = {1},
  journal = {Scientific Reports},
  shortjournal = {Sci Rep},
  author = {da Costa, Danny and Kornemann, Lukas and Goebel, Rainer and Senden, Mario},
  urldate = {2024-08-08},
  year = {2024},
  langid = {english},
  note = {Publisher: Nature Publishing Group},
}

@article{golomb_native_2008,
  title = {The Native Coordinate System of Spatial Attention Is Retinotopic},
  volume = {28},
  rightsholder = {https://creativecommons.org/licenses/by-nc-sa/4.0/},
  issn = {0270-6474, 1529-2401},
  url = {https://www.jneurosci.org/lookup/doi/10.1523/JNEUROSCI.2525-08.2008},
  doi = {10.1523/JNEUROSCI.2525-08.2008},
  pages = {10654--10662},
  number = {42},
  journal = {The Journal of Neuroscience},
  shortjournal = {J. Neurosci.},
  author = {Golomb, Julie D. and Chun, Marvin M. and Mazer, James A.},
  urldate = {2024-08-30},
  year = {2008},
  langid = {english},
}

@article{nauhaus_efficient_2016,
  title = {Efficient Receptive Field Tiling in Primate V1},
  volume = {91},
  issn = {08966273},
  url = {https://linkinghub.elsevier.com/retrieve/pii/S089662731630366X},
  doi = {10.1016/j.neuron.2016.07.015},
  pages = {893--904},
  number = {4},
  journal = {Neuron},
  shortjournal = {Neuron},
  author = {Nauhaus, Ian and Nielsen, Kristina J. and Callaway, Edward M.},
  urldate = {2024-08-30},
  year = {2016},
  langid = {english},
}

@article{Lukanov2021,
  title = {Biologically {{Inspired Deep Learning Model}} for {{Efficient Foveal-Peripheral Vision}}},
  author = {Lukanov, Hristofor and König, Peter and Pipa, Gordon},
  year = {2021},
  journal = {Frontiers in Computational Neuroscience},
  volume = {15},
  issn = {1662-5188},
  doi = {10.3389/fncom.2021.746204},
  url = {https://www.frontiersin.org/article/10.3389/fncom.2021.746204},
  urldate = {2022-06-13}
}

@article{Dauce2020,
  title = {A Dual Foveal-Peripheral Visual Processing Model Implements Efficient Saccade Selection},
  author = {Daucé, Emmanuel and Albiges, Pierre and Perrinet, Laurent U.},
  year = {2020},
  journal = {Journal of Vision},
  shortjournal = {Journal of Vision},
  volume = {20},
  number = {8},
  pages = {22--22},
  publisher = {{The Association for Research in Vision and Ophthalmology}},
  issn = {1534-7362},
  doi = {10.1167/jov.20.8.22},
  url = {https://jov.arvojournals.org/article.aspx?articleid=2770680},
  urldate = {2021-07-02}
}

@inproceedings{Dauce2020a,
  ids = {Dauce2020ai},
  title = {Visual {{Search}} as {{Active Inference}}},
  booktitle = {Active {{Inference}}},
  author = {Daucé, Emmanuel and Perrinet, Laurent},
  editor = {Verbelen, Tim and Lanillos, Pablo and Buckley, Christopher L. and De Boom, Cedric},
  year = {2020},
  series = {Communications in {{Computer}} and {{Information Science}}},
  pages = {165--178},
  publisher = {Springer International Publishing},
  location = {Cham},
  doi = {10.1007/978-3-030-64919-7_17},
  isbn = {978-3-030-64919-7}
}

@article{He2015wrn,
  ids = {he2016deep},
  title = {Deep {{Residual Learning}} for {{Image Recognition}}},
  author = {He, Kaiming and Zhang, Xiangyu and Ren, Shaoqing and Sun, Jian},
  year = {2015},
  journal = {arXiv:1512.03385 [cs.CV]},
  eprint = {1512.03385},
  eprinttype = {arXiv},
  eprintclass = {cs.CV},
  doi = {10.1109/CVPR.2016.90},
  url = {http://arxiv.org/abs/1512.03385},
  urldate = {2023-07-20}
}

@article{Jeremie2023,
  title = {Ultrafast {{Image Categorization}} in {{Biology}} and {{Neural Models}}},
  author = {Jérémie, Jean-Nicolas and Perrinet, Laurent U.},
  year = {2023},
  journal = {Vision},
  shortjournal = {Vision (Basel)},
  volume = {7},
  number = {2},
  pages = {29},
  issn = {2411-5150},
  doi = {10.3390/vision7020029}
}

@article{Potter1975,
  title = {Meaning in Visual Search},
  author = {Potter, Mary C.},
  year = {1975},
  journal = {Science},
  volume = {187},
  number = {4180},
  pages = {965--966},
  publisher = {American Association for the Advancement of Science},
  doi = {10.1126/science.1145183},
  url = {https://www.science.org/doi/10.1126/science.1145183},
  urldate = {2024-01-16}
}

@article{Hao2021,
  title = {Retina-like Imaging and Its Applications: {A} Brief Review},
  author = {Hao, Qun and Tao, Yu and Cao, Jie and Tang, Mingyuan and Cheng, Yang and Zhou, Dong and Ning, Yaqian and Bao, Chun and Cui, Huan},
  year = {2021},
  journal = {Applied Sciences},
  shortjournal = {Applied Sciences},
  volume = {11},
  number = {15},
  pages = {7058},
  issn = {2076-3417},
  doi = {10.3390/app11157058},
  url = {https://www.mdpi.com/2076-3417/11/15/7058},
  urldate = {2024-01-08}
}

@inproceedings{Palander2008,
  title = {Epipolar Geometry and Log-Polar Transform in Wide Baseline Stereo Matching},
  booktitle = {2008 19th {{International Conference}} on {{Pattern Recognition}}},
  author = {Palander, Kimmo and Brandt, Sami S.},
  year = {2008},
  month = dec,
  pages = {1--4},
  publisher = {IEEE},
  location = {Tampa, FL, USA},
  issn = {1051-4651},
  doi = {10.1109/ICPR.2008.4761515},
  url = {http://ieeexplore.ieee.org/document/4761515/},
  urldate = {2023-12-20}
}

@article{JavierTraver2010,
  ids = {Traver-Roig2010},
  title = {A Review of Log-Polar Imaging for Visual Perception in Robotics},
  author = {Javier Traver, V. and Bernardino, Alexandre},
  year = {2010},
  journal = {Robotics and Autonomous Systems},
  shortjournal = {Robotics and Autonomous Systems},
  volume = {58},
  number = {4},
  pages = {378--398},
  publisher = {Elsevier},
  issn = {09218890},
  doi = {10.1016/j.robot.2009.10.002},
  url = {https://linkinghub.elsevier.com/retrieve/pii/S0921889009001687},
  urldate = {2023-12-20}
}

@article{Cao2021,
  title = {{{LPNet}}: {{Retina Inspired Neural Network}} for {{Object Detection}} and {{Recognition}}},
  shorttitle = {{{LPNet}}},
  author = {Cao, Jie and Bao, Chun and Hao, Qun and Cheng, Yang and Chen, Chenglin},
  year = {2021},
  journal = {Electronics},
  volume = {10},
  number = {22},
  pages = {2883},
  publisher = {Multidisciplinary Digital Publishing Institute},
  issn = {2079-9292},
  doi = {10.3390/electronics10222883},
  url = {https://www.mdpi.com/2079-9292/10/22/2883},
  urldate = {2023-11-16},
  issue = {22}
}

@unpublished{Remmelzwaal2019,
  title = {Human Eye Inspired Log-Polar Pre-Processing for Neural Networks},
  author = {Remmelzwaal, Leendert A. and Mishra, Amit and Ellis, George F. R.},
  year = {2019},
  month = nov,
  eprint = {1911.01141},
  eprinttype = {arXiv},
  eprintclass = {cs},
  doi = {10.48550/arXiv.1911.01141},
  url = {http://arxiv.org/abs/1911.01141},
  urldate = {2023-11-16},
  pubstate = {prepublished}
}

@unpublished{Huang2017,
  title = {Adversarial {{Attacks}} on {{Neural Network Policies}}},
  author = {Huang, Sandy and Papernot, Nicolas and Goodfellow, Ian and Duan, Yan and Abbeel, Pieter},
  year = {2017},
  month = feb,
  eprint = {1702.02284},
  eprinttype = {arXiv},
  eprintclass = {cs, stat},
  url = {http://arxiv.org/abs/1702.02284},
  urldate = {2023-09-28},
  pubstate = {prepublished}
}

@article{Guyonneau2006,
  title = {Animals Roll around the Clock: {{The}} Rotation Invariance of Ultrarapid Visual Processing},
  shorttitle = {Animals Roll around the Clock},
  author = {Guyonneau, Rudy and Kirchner, Holle and Thorpe, Simon J.},
  year = {2006},
  month = sep,
  journal = {Journal of Vision},
  shortjournal = {Journal of Vision},
  volume = {6},
  number = {10},
  pages = {1},
  issn = {1534-7362},
  doi = {10.1167/6.10.1},
  url = {https://doi.org/10.1167/6.10.1},
  urldate = {2024-01-23}
}

@unpublished{Zhang2024,
  title = {Opti-{{CAM}}: {{Optimizing}} Saliency Maps for Interpretability},
  shorttitle = {Opti-{{CAM}}},
  author = {Zhang, Hanwei and Torres, Felipe and Sicre, Ronan and Avrithis, Yannis and Ayache, Stephane},
  year = {2024},
  month = feb,
  eprint = {2301.07002},
  eprinttype = {arXiv},
  eprintclass = {cs},
  publisher = {arXiv},
  url = {http://arxiv.org/abs/2301.07002},
  urldate = {2024-03-11}
}

@inproceedings{amorim_analysing_2018,
  title = {Analysing rotation-invariance of a log-polar transformation in convolutional neural networks},
  url = {https://ieeexplore.ieee.org/abstract/document/8489295},
  doi = {10.1109/IJCNN.2018.8489295},
  eventtitle = {2018 International Joint Conference on Neural Networks ({IJCNN})},
  pages = {1--6},
  booktitle = {2018 International Joint Conference on Neural Networks ({IJCNN})},
  author = {Amorim, Marta and Bortoloti, Frederico and Ciarelli, Patrick Marques and de Oliveira, Elias and de Souza, Alberto Ferreira},
  urldate = {2025-09-17},
  date = {2018-07},
  note = {{ISSN}: 2161-4407},
}

@article{han_scale_2020,
  title = {Scale and translation-invariance for novel objects in human vision},
  volume = {10},
  issn = {2045-2322},
  url = {https://www.nature.com/articles/s41598-019-57261-6},
  doi = {10.1038/s41598-019-57261-6},
  pages = {1411},
  number = {1},
  journaltitle = {Scientific Reports},
  shortjournal = {Sci Rep},
  author = {Han, Yena and Roig, Gemma and Geiger, Gad and Poggio, Tomaso},
  urldate = {2025-09-17},
  date = {2020-01-29},
  langid = {english},
}

@article{pun_log-polar_2003,
  title = {Log-polar wavelet energy signatures for rotation and scale invariant texture classification},
  volume = {25},
  issn = {1939-3539},
  url = {https://ieeexplore.ieee.org/abstract/document/1195993},
  doi = {10.1109/TPAMI.2003.1195993},
  pages = {590--603},
  number = {5},
  journaltitle = {{IEEE} Transactions on Pattern Analysis and Machine Intelligence},
  author = {Pun, Chi-Man and Lee, Moon-Chuen},
  urldate = {2025-09-17},
  date = {2003-05},
}

@unpublished{Wang2020,
  title = {Score-{{CAM}}: {{Score-Weighted Visual Explanations}} for {{Convolutional Neural Networks}}},
  shorttitle = {Score-{{CAM}}},
  author = {Wang, Haofan and Wang, Zifan and Du, Mengnan and Yang, Fan and Zhang, Zijian and Ding, Sirui and Mardziel, Piotr and Hu, Xia},
  year = {2020},
  month = apr,
  eprint = {1910.01279},
  eprinttype = {arXiv},
  eprintclass = {cs},
  publisher = {arXiv},
  url = {http://arxiv.org/abs/1910.01279},
  urldate = {2024-03-11}
}

@article{Selvaraju2020,
  title = {Grad-{{CAM}}: {{Visual Explanations}} from {{Deep Networks}} via {{Gradient-based Localization}}},
  shorttitle = {Grad-{{CAM}}},
  author = {Selvaraju, Ramprasaath R. and Cogswell, Michael and Das, Abhishek and Vedantam, Ramakrishna and Parikh, Devi and Batra, Dhruv},
  year = {2020},
  month = feb,
  journal = {International Journal of Computer Vision},
  shortjournal = {Int J Comput Vis},
  volume = {128},
  number = {2},
  eprint = {1610.02391},
  eprinttype = {arXiv},
  eprintclass = {cs},
  pages = {336--359},
  issn = {0920-5691, 1573-1405},
  doi = {10.1007/s11263-019-01228-7},
  url = {http://arxiv.org/abs/1610.02391},
}

@article{Tootell1998,
  title = {The {{Retinotopy}} of {{Visual Spatial Attention}}},
  author = {Tootell, Roger B.H and Hadjikhani, Nouchine and Hall, E.Kevin and Marrett, Sean and Vanduffel, Wim and Vaughan, J.Thomas and Dale, Anders M},
  year = {1998},
  month = dec,
  journal = {Neuron},
  shortjournal = {Neuron},
  volume = {21},
  number = {6},
  pages = {1409--1422},
  issn = {08966273},
  doi = {10.1016/S0896-6273(00)80659-5},
  url = {https://linkinghub.elsevier.com/retrieve/pii/S0896-6273(00)80659-5},
  urldate = {2024-04-23}
}

@unpublished{Kim2020,
  title = {{{CyCNN}}: A Rotation Invariant {{CNN}} Using {{Polar Mapping}} and {{Cylindrical Convolution Layers}}},
  shorttitle = {{{CyCNN}}},
  author = {Kim, Jinpyo and Jung, Wooekun and Kim, Hyungmo and Lee, Jaejin},
  year = {2020},
  month = jul,
  eprint = {2007.10588},
  eprinttype = {arXiv},
  eprintclass = {cs, eess},
  url = {http://arxiv.org/abs/2007.10588},
  urldate = {2024-04-23},
  pubstate = {prepublished}
}

@misc{esteves_learning_2018,
  title = {Learning {SO}(3) Equivariant Representations with Spherical {CNNs}},
  url = {http://arxiv.org/abs/1711.06721},
  doi = {10.48550/arXiv.1711.06721},
  number = {{arXiv}:1711.06721},
  publisher = {{arXiv}},
  author = {Esteves, Carlos and Allen-Blanchette, Christine and Makadia, Ameesh and Daniilidis, Kostas},
  urldate = {2025-09-10},
  date = {2018-09-28},
  langid = {english},
  eprinttype = {arxiv},
  eprint = {1711.06721 [cs]},
}

@article{Dougherty2003,
  title = {Visual Field Representations and Locations of Visual Areas V1/2/3 in Human Visual Cortex},
  author = {Dougherty, Robert F. and Koch, Volker M. and Brewer, Alyssa A. and Fischer, Bernd and Modersitzki, Jan and Wandell, Brian A.},
  year = {2003},
  month = oct,
  journal = {Journal of Vision},
  shortjournal = {Journal of Vision},
  volume = {3},
  number = {10},
  pages = {1},
  issn = {1534-7362},
  doi = {10.1167/3.10.1},
  url = {http://jov.arvojournals.org/article.aspx?doi=10.1167/3.10.1},
  urldate = {2024-04-23}
}

@article{Sandini1980,
  title = {An Anthropomorphic Retina-like Structure for Scene Analysis},
  author = {Sandini, Giulio and Tagliasco, Vincenzo},
  year = {1980},
  month = dec,
  journal = {Computer Graphics and Image Processing},
  shortjournal = {Computer Graphics and Image Processing},
  volume = {14},
  number = {4},
  pages = {365--372},
  issn = {0146664X},
  doi = {10.1016/0146-664X(80)90026-X},
  url = {https://linkinghub.elsevier.com/retrieve/pii/0146664X8090026X},
  urldate = {2024-04-24}
}

@article{Anstis1974,
  title = {A Chart Demonstrating Variations in Acuity with Retinal Position},
  author = {Anstis, S.M.},
  year = {1974},
  month = jul,
  journal = {Vision Research},
  shortjournal = {Vision Research},
  volume = {14},
  number = {7},
  pages = {589--592},
  issn = {0042-6989},
  doi = {10.1016/0042-6989(74)90049-2},
  url = {https://linkinghub.elsevier.com/retrieve/pii/0042698974900492},
  urldate = {2024-04-24}
}

@unpublished{Kubilius2018,
  title = {{{CORnet}}: {{Modeling}} the {{Neural Mechanisms}} of {{Core Object Recognition}}},
  shorttitle = {{{CORnet}}},
  author = {Kubilius, Jonas and Schrimpf, Martin and Nayebi, Aran and Bear, Daniel and Yamins, Daniel L. K. and DiCarlo, James J.},
  year = {2018},
  month = sep,
  doi = {10.1101/408385},
  url = {http://biorxiv.org/lookup/doi/10.1101/408385},
  urldate = {2024-04-25},
  pubstate = {prepublished}
}

@article{Rothkopf2016,
  ids = {rothkopf_task_2016},
  title = {Task and Context Determine Where You Look},
  author = {Rothkopf, Constantin A. and Ballard, Dana H. and Hayhoe, Mary M.},
  year = {2016},
  month = jul,
  journal = {Journal of Vision},
  shortjournal = {Journal of Vision},
  volume = {7},
  number = {14},
  pages = {16},
  issn = {1534-7362},
  doi = {10.1167/7.14.16},
  url = {https://doi.org/10.1167/7.14.16},
  urldate = {2024-07-05}
}

@unpublished{Mineault2021,
  title = {Your Head Is There to Move You Around: {{Goal-driven}} Models of the {{Primate Dorsal Pathway}}},
  shorttitle = {Your Head Is There to Move You Around},
  author = {Mineault, Patrick J and Bakhtiari, Shahab and Richards, Blake A and Pack, Christopher C},
  year = {2021},
  month = jul,
  journal = {biorxiv},
  doi = {10.1101/2021.07.09.451701},
  url = {http://biorxiv.org/lookup/doi/10.1101/2021.07.09.451701},
  urldate = {2024-02-06}
}

@unpublished{Bakhtiari2021,
  title = {The Functional Specialization of Visual Cortex Emerges from Training Parallel Pathways with Self-Supervised Predictive Learning},
  author = {Bakhtiari, Shahab and Mineault, Patrick and Lillicrap, Tim and Pack, Christopher C. and Richards, Blake},
  year = {2021},
  month = jun,
  journal = {biorxiv},
  doi = {10.1101/2021.06.18.448989},
  url = {http://biorxiv.org/lookup/doi/10.1101/2021.06.18.448989},
  urldate = {2024-02-06}
}

@article{Jaderberg2016,
  title = {Spatial {{Transformer Networks}}},
  author = {Jaderberg, Max and Simonyan, Karen and Zisserman, Andrew and Kavukcuoglu, Koray},
  year = {2016},
  month = feb,
  journal = {arXiv},
  doi = {10.48550/arXiv.1506.02025},
  url = {http://arxiv.org/abs/1506.02025},
  urldate = {2023-09-29}
}

@unpublished{szegedy2013intriguing,
  title = {Intriguing Properties of Neural Networks},
  author = {Szegedy, Christian and Zaremba, Wojciech and Sutskever, Ilya and Bruna, Joan and Erhan, Dumitru and Goodfellow, Ian and Fergus, Rob},
  year = {2013},
  eprint = {1312.6199},
  eprinttype = {arXiv}
}

@inproceedings{Lempitsky2018,
  title = {Deep {{Image Prior}}},
  booktitle = {2018 {{IEEE}}/{{CVF Conference}} on {{Computer Vision}} and {{Pattern Recognition}}},
  author = {Lempitsky, Victor and Vedaldi, Andrea and Ulyanov, Dmitry},
  year = {2018},
  month = jun,
  pages = {9446--9454},
  publisher = {IEEE},
  location = {Salt Lake City, UT},
  doi = {10.1109/CVPR.2018.00984},
  url = {https://ieeexplore.ieee.org/document/8579082/},
  urldate = {2023-09-27},
  isbn = {978-1-5386-6420-9},
}

@article{Weinberg1997,
  title = {Are {{Topographic Maps Fundamental}} to {{Sensory Processing}}?},
  author = {Weinberg, Richard J},
  year = {1997},
  month = jan,
  journal = {Brain Research Bulletin},
  volume = {44},
  number = {2},
  pages = {113--116},
  issn = {0361-9230},
  doi = {10.1016/S0361-9230(97)00095-6},
  url = {https://www.sciencedirect.com/science/article/pii/S0361923097000956},
  urldate = {2023-09-27}
}

@article{Yarbus1961,
  title = {Eye Movements during the Examination of Complicated Objects},
  author = {Yarbus, A},
  year = {1961},
  journal = {Biofizika},
  volume = {6(2)},
  eprint = {14040367},
  eprinttype = {pmid},
  pages = {52--56},
  issn = {0006-3029}
}

@article{Antonelli2015,
  title = {Speeding up the Log-Polar Transform with Inexpensive Parallel Hardware: Graphics Units and Multi-Core Architectures},
  author = {Antonelli, Marco and Igual, Francisco D. and Ramos, Francisco and Traver, V. Javier},
  year = {2015},
  journal = {Journal of Real-Time Image Processing},
  volume = {10},
  number = {3},
  pages = {533--550},
  issn = {18618200},
  doi = {10.1007/s11554-012-0281-6},
  url = {https://doi.org/10.1007/s11554-012-0281-6}
}

@article{Crouzet2011,
  title = {What Are the Visual Features Underlying Rapid Object Recognition?},
  author = {Crouzet, Sébastien M.},
  year = {2011},
  journal = {Frontiers in Psychology},
  volume = {2},
  doi = {10.3389/fpsyg.2011.00326}
}

@inproceedings{dabane2022you,
  title = {What You See Is What You Transform: {{Foveated}} Spatial Transformers as a Bio-Inspired Attention Mechanism},
  booktitle = {2022 International Joint Conference on Neural Networks ({{IJCNN}})},
  author = {Dabane, Ghassan and Perrinet, Laurent U and Daucé, Emmanuel},
  year = {2022},
  pages = {1--8},
  publisher = {IEEE},
  doi = {10.1109/IJCNN55064.2022.9892313}
}

@article{Maiello2020,
  title = {Near-Optimal Combination of Disparity across a Log-Polar Scaled Visual Field},
  author = {Maiello, Guido and Chessa, Manuela and Bex, Peter J and Solari, Fabio},
  year = {2020},
  journal = {PLoS Computational Biology},
  volume = {16},
  number = {4},
  pages = {e1007699},
  publisher = {Public Library of Science San Francisco, CA USA},
  doi = {10.1371/journal.pcbi.1007699}
}

@incollection{Paszke2019,
  title = {{{PyTorch}}: {{An}} Imperative Style, High-Performance Deep Learning Library},
  booktitle = {Advances in Neural Information Processing Systems 32},
  author = {Paszke, Adam and Gross, Sam and Massa, Francisco and Lerer, Adam and Bradbury, James and Chanan, Gregory and Killeen, Trevor and Lin, Zeming and Gimelshein, Natalia and Antiga, Luca and Desmaison, Alban and Kopf, Andreas and Yang, Edward and DeVito, Zachary and Raison, Martin and Tejani, Alykhan and Chilamkurthy, Sasank and Steiner, Benoit and Fang, Lu and Bai, Junjie and Chintala, Soumith},
  editor = {Wallach, H. and Larochelle, H. and Beygelzimer, A. and Fox, E. and Garnett, R.},
  year = {2019},
  pages = {8024--8035},
  publisher = {Curran Associates, Inc.}
}

@book{Polyak1941,
	address = {University of Chicago Press},
	title = {The retina: the anatomy and the histology of the retina in man, ape, and monkey, including the consideration of visual functions, the history of physiological optics, and the histological laboratory technique},
	shorttitle = {The retina},
	publisher = {Chicago},
	author = {Polyak, S.L.},
	year = {1941},
}

@article{Rousselet2003,
  title = {Is It an Animal? {{Is}} It a Human Face? {{Fast}} Processing in Upright and Inverted Natural Scenes.},
  author = {Rousselet, G. A. and Macé, M. J.-M. and Fabre-Thorpe, Michèle},
  year = {2003},
  journal = {Journal of Vision},
  volume = {3},
  pages = {440--455},
  doi = {10.1167/3.6.5}
}

@article{Russakovsky2015,
  title = {{{ImageNet}} Large Scale Visual Recognition Challenge},
  author = {Russakovsky, Olga and Deng, Jia and Su, Hao and Krause, Jonathan and Satheesh, Sanjeev and Ma, Sean and Huang, Zhiheng and Karpathy, Andrej and Khosla, Aditya and Bernstein, Michael and Berg, Alexander C. and Fei-Fei, Li},
  year = {2015},
  journal = {International Journal of Computer Vision (IJCV)},
  volume = {115},
  pages = {211--252},
  doi = {10.1007/s11263-015-0816-y}
}

@unpublished{Simonyan2015,
  title = {Very {{Deep Convolutional Networks}} for {{Large-Scale Image Recognition}}},
  author = {Simonyan, Karen and Zisserman, Andrew},
  year = {2015},
  eprint = {1409.1556},
  eprinttype = {arXiv},
  eprintclass = {cs}
}

@misc{Yu2021,
  title = {{{AP-10K}}: A Benchmark for Animal Pose Estimation in the Wild},
  author = {Yu, Hang and Xu, Yufei and Zhang, Jing and Zhao, Wei and Guan, Ziyu and Tao, Dacheng},
  year = {2021},
  eprint = {2108.12617},
  eprinttype = {arXiv},
  eprintclass = {cs.CV}
}

@article{lewis2004understanding,
  title = {Understanding Cone Distributions from Saccadic Dynamics. {{Is}} Information Rate Maximised?},
  author = {Lewis, Alex and Garcia, Raquel and Zhaoping, Li},
  year = {2004},
  journal = {Neurocomputing},
  volume = {58},
  pages = {807--813},
  publisher = {Elsevier},
  doi = {10.1016/j.neucom.2004.01.131}
}

@article{Collin2018,
  title = {Scene through the Eyes of an Apex Predator: A Comparative Analysis of the Shark Visual System},
  shorttitle = {Scene through the Eyes of an Apex Predator},
  author = {Collin, Shaun P},
  year = {2018},
  journal = {Clinical and Experimental Optometry},
  volume = {101},
  number = {5},
  pages = {624--640},
  issn = {1444-0938},
  doi = {10.1111/cxo.12823},
  url = {https://onlinelibrary.wiley.com/doi/abs/10.1111/cxo.12823},
  urldate = {2024-08-08}
}

@incollection{Mitkus2018,
  title = {Raptor {{Vision}}},
  booktitle = {Oxford {{Research Encyclopedia}} of {{Neuroscience}}},
  author = {Mitkus, Mindaugas and Potier, Simon and Martin, Graham R. and Duriez, Olivier and Kelber, Almut},
  year = {2018},
  month = apr,
  doi = {10.1093/acrefore/9780190264086.013.232},
  url = {https://oxfordre.com/neuroscience/display/10.1093/acrefore/9780190264086.001.0001/acrefore-9780190264086-e-232?ref=PDF},
  urldate = {2024-08-08},
  isbn = {978-0-19-026408-6}
}

@article{najemnik2005optimal,
  title = {Optimal Eye Movement Strategies in Visual Search},
  author = {Najemnik, Jiri and Geisler, Wilson S},
  year = {2005},
  journal = {Nature},
  volume = {434},
  number = {7031},
  pages = {387--391},
  publisher = {Nature Publishing Group UK London},
  doi = {10.1038/nature03390}
}

@article{dauce2018active,
  title = {Active Fovea-Based Vision through Computationally-Effective Model-Based Prediction},
  author = {Daucé, Emmanuel},
  year = {2018},
  journal = {Frontiers in neurorobotics},
  volume = {12},
  pages = {76},
  publisher = {Frontiers Media SA},
  doi = {10.3389/fnbot.2018.00076}
}

@article{lewis2003distribution,
  title = {The Distribution of Visual Objects on the Retina: Connecting Eye Movements and Cone Distributions},
  author = {Lewis, Alex and Garcia, Raquel and Zhaoping, Li},
  year = {2003},
  journal = {Journal of vision},
  volume = {3},
  number = {11},
  pages = {21--21},
  publisher = {{The Association for Research in Vision and Ophthalmology}},
  doi = {10.1167/3.11.21}
}

@article{noton_scanpaths_1971,
  title = {Scanpaths in Eye Movements during Pattern Perception},
  author = {Noton, David and Stark, Lawrence},
  year = {1971},
  month = jan,
  journal = {Science},
  shortjournal = {Science},
  volume = {171},
  number = {3968},
  pages = {308--311},
  issn = {0036-8075, 1095-9203},
  doi = {10.1126/science.171.3968.308},
  url = {https://www.science.org/doi/10.1126/science.171.3968.308},
  urldate = {2024-04-24}
}

@article{itti_koch2005,
	title = {Computational modelling of visual attention},
	volume = {2},
	copyright = {2001 Springer Nature Limited},
	issn = {1471-0048},
	url = {https://www.nature.com/articles/35058500},
	doi = {10.1038/35058500},
	language = {en},
	number = {3},
	urldate = {2024-11-05},
	journal = {Nature Reviews Neuroscience},
	author = {Itti, Laurent and Koch, Christof},
	month = mar,
	year = {2001},
	note = {Publisher: Nature Publishing Group},
	pages = {194--203},
}

@article{wolfe1994guided,
  title = {Guided Search 2.0 a Revised Model of Visual Search},
  author = {Wolfe, Jeremy M},
  year = {1994},
  journal = {Psychonomic bulletin \& review},
  volume = {1},
  pages = {202--238},
  publisher = {Springer},
  doi = {10.3758/BF03200774}
}

\end{document}